\documentclass[final,5p,times,twocolumn]{elsarticle}

\usepackage{amssymb}
\usepackage{amsmath}
\usepackage{booktabs}
\usepackage{graphicx}
\usepackage{makecell}
\usepackage{hyperref}
\usepackage{adjustbox}
\usepackage{multirow}
\usepackage{nicefrac}
\usepackage[monochrome]{xcolor} 
\usepackage{floatrow}

\usepackage[utf8]{inputenc}

\biboptions{sort&compress}

\newfloatcommand{capbtabbox}{table}[][\FBwidth]

\DeclareMathOperator*{\argmax}{argmax}

\journal{}

\begin{document}

\begin{frontmatter}

\title{Dense Center-Direction Regression for Object Counting and Localization with Point Supervision}

\author[FRI]{Domen Tabernik}
\ead{domen.tabernik@fri.uni-lj.si}

\author[FRI]{Jon Muhovi\v{c}}
\ead{jon.muhovic@fri.uni-lj.si}

\author[FRI]{Danijel Skočaj}
\ead{danijel.skocaj@fri.uni-lj.si}

\affiliation[FRI]{organization={Faculty of Computer and Information Science, University of Ljubljana},
            addressline={Vecna pot 113}, 
            city={Ljubljana},
            country={Slovenia}}

\begin{abstract}
Object counting and localization problems are commonly addressed with point supervised learning, which allows the use of less labor-intensive point annotations. However, learning based on point annotations poses challenges due to the high imbalance between the sets of annotated and unannotated pixels, which is often treated with Gaussian smoothing of point annotations and focal loss. However, these approaches still focus on the pixels in the immediate vicinity of the point annotations and exploit the rest of the data only indirectly. In this work, we propose a novel approach termed \emph{CeDiRNet} for point-supervised learning that uses a dense regression of directions pointing towards the nearest object centers, i.e. \emph{center-directions}. This provides greater support for each center point arising from many surrounding pixels pointing towards the object center. We propose a formulation of center-directions that allows the problem to be split into the domain-specific dense regression of center-directions and the final localization task based on a small, lightweight, and domain-agnostic localization network that can be trained with synthetic data completely independent of the target domain. We demonstrate the performance of the proposed method on six different datasets for object counting and localization, and show that it outperforms the existing state-of-the-art methods. {The code is accessible on GitHub at~\url{https://github.com/vicoslab/CeDiRNet.git}.}

\end{abstract}


\end{frontmatter}

\section{Introduction}

Object counting is a computer vision task with the goal of counting the number of visual objects of a specific category in visual data. Counting is often addressed with a direct regression of the count number~\cite{Chan2008} or the density estimation map~\cite{Liu2018,Rong2021,Wan2020,Lei2021}. However, returning only the count or a density map is not sufficient for many applications where information about the location of the object is also required for additional verification and explanation, such as industrial quality control applications or in some remote sensing domains (counting vehicles, buildings, or ships). In this work we focus on the problem of simultaneous counting and localization of objects from visual data. 

One solution would be to leverage existing state-of-the-art generic object detectors, e.g., Mask R-CNN~\cite{He2017}, YOLO~\cite{Redmon2016}, FCOS~\cite{Tian2019}, or AccLoc~\cite{Piao2022}, and count the number of detections. However, generic detectors focus on finding accurate boundaries of objects, which is not required for counting applications, and thus use too much model capacity to achieve good boundary accuracy instead of focusing on counting accuracy. They are also prone to errors in densely packed scenes~\cite{Goldman2019} due to overlap of bounding boxes, while they also require detailed annotations (masks or bounding boxes) that are not required for accurate counting and localization. For counting, localization with a single point is sufficient.

\begin{figure}
    \centering
    \includegraphics[width=0.95\linewidth]{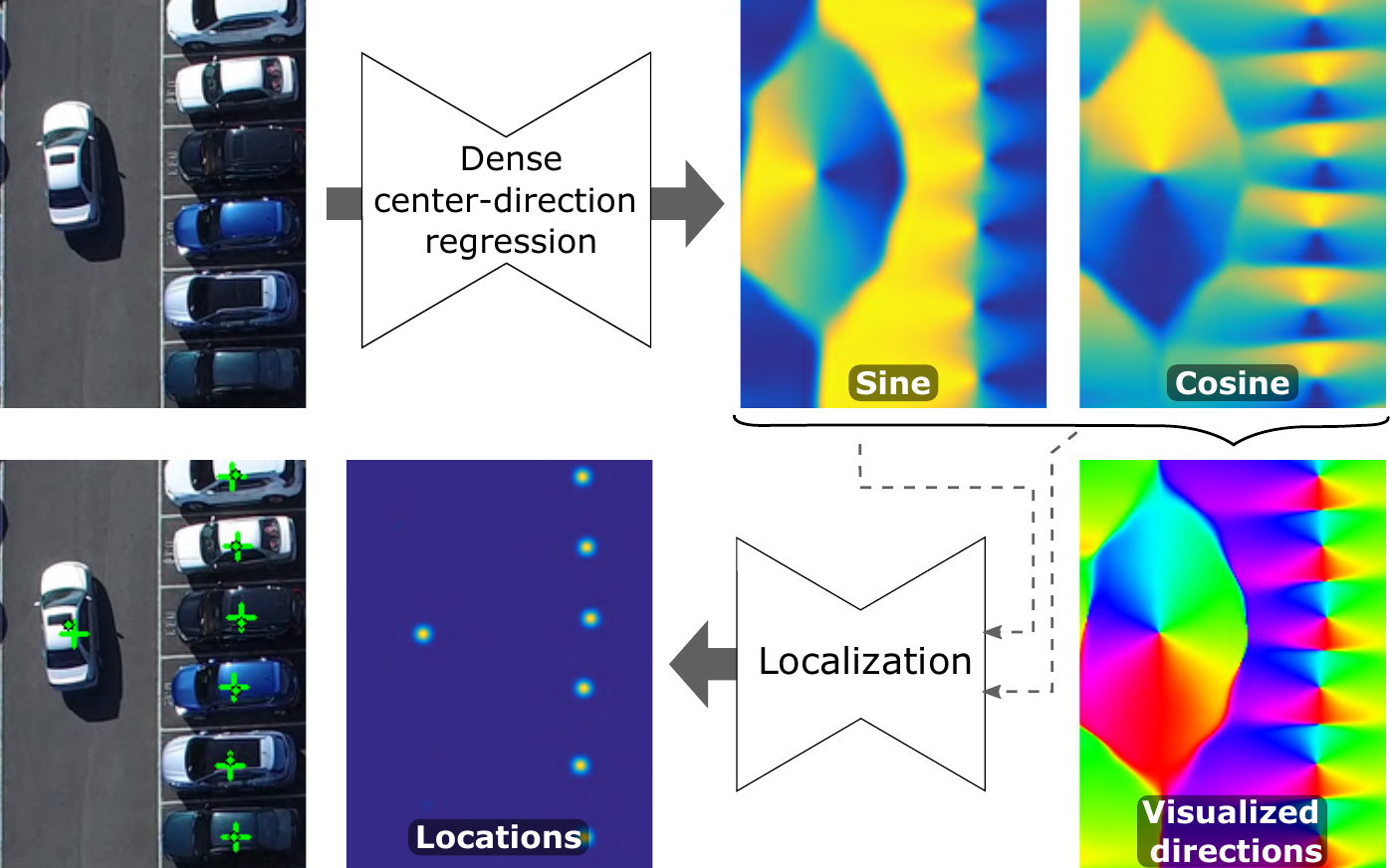}
    \caption{Overview of the proposed CeDiRNet approach.}
    \label{fig:intro}
\end{figure}

Moreover, using only point labels makes the annotation process significantly less labor intensive compared to detailed annotation with bounding boxes. Studies comparing annotation effort have found that annotation with center points is 77\% faster than annotation with bounding boxes, and more than 3.5 times faster than annotation with four-point polygons. This becomes even more important when there are a large number of objects in the image, as is the case in the counting problem. For this reason, recent approaches to counting and localization ~\cite{Ribera2019, Wang2021, Tong2021} employ methods based on point-supervision that predict the object location as a single point. This reduces the amount of annotation required and also leads to the replacement of bounding box location regression or segmentation mask estimation with object center probability prediction.

In most such approaches~\cite{Wang2021, Tong2021}, the prediction of an object's center is done by directly learning the probability for every pixel in an image that it belongs to the object center. This can be implemented as a binary segmentation problem using cross-entropy loss, with only a single object center point marked as foreground. However, this also leads to an extremely unbalanced problem due to a large number of background pixels compared to the number of foreground pixels. Most approaches address this problem with focal loss~\cite{Lin2017a} and with Gaussian blur to spread the center point and expand the foreground region, while some~\cite{Ribera2019} avoid this problem by replacing the cross entropy loss with the weighted Hausdorff distance. {Nevertheless, these methods still use direct regression of a per-pixel likelihood or saliency map of the center point thus regressing to zero values in the background. However, regression to zero values away from the center does not encourage the model to find useful information farther away from the center, which may still be useful for predicting the center, especially if the object is less prominent in inference time than during training. Some information may be found, but this is not explicitly encouraged by the loss. Adding regression of direct offset values to the centers from the background location would address this problem, but such methods do not regress the values outside the centers~\citep{Zhou2019} or outside the instance mask~\citep{Neven2019}.
}

{
Moreover, when regressing the offsets to centers with gradient descent methods, this results in a small gradient for the former and a large gradient for the latter, creating an imbalance between them during training. Effectively, offsets that are farther away suppress learning of offsets that are closer to the center, even though the closer offsets tend to be more important. Also, direct offset values are not invariant to scale changes and the network must regress different values depending on image size or object scale. In CenterNet~\cite{Zhou2019}, this problem is avoided by learning regressions only in locations around the centers, while \cite{Neven2019,Zhou2021} add a scale factor to reduce the size of the vectors. Neven et al.~\cite{Neven2019} also performs the regression only within the instance mask, which mitigates this problem. Current methods that regress direct offsets from all surrounding pixels also require complex post-processing to infer centers from the offsets, such as clustering~\cite{Neven2019} or Hough voting~\cite{Zhou2021}, making them unsuitable for locating a large number of objects common in counting.
}

We propose a novel point-supervision approach for object counting and localization termed \emph{CeDiRNet}. The proposed approach densely regresses directions pointing to the nearest object center, rather than relying on a direct regression of the center's probability map or direct offsets. We refer to such regressed directions as \emph{center-directions}. Regressing the center-direction for all pixel locations in the image encourages the model to find features that are correlated with the center, even if they are farther from the center, and thus providing greater support for the center from a larger surrounding area. {Center-directions can be viewed as normalized offsets that, unlike direct offsets, are invariant to changes in scale (the network can learn the same value regardless of image or object size) while also avoiding the unbalanced learning issues between the values that are near and those that are farther from the center that occur with direct offset regression.} We also formulate the center-directions such that they provide a specific signature at the center of the object (see Fig.~\ref{fig:intro}), making it possible to find the object center with a small, lightweight localization network applied to the regressed center-direction thus avoiding complex post-processing method such as clustering or Hough voting. Moreover, the appearance of the center-directions is not domain-specific, but is similar for different categories, allowing the use of a domain-agnostic localization network that can be trained with synthetically generated input data completely independent of a specific domain.

The contribution of the proposed approach is threefold: (a) the dense prediction of the center-directions allows for large support from pixels farther away from the center, (b) only the center-direction regression is trained on the target domain, while the localization network is fully domain-agnostic, trained once using synthetic data and shared among all domains, and (c) point-supervision reduces the annotation effort. We demonstrate the benefits with an extensive evaluation on six different datasets, showing that our approach is able to outperform existing state-of-the-art approaches for counting and localization. Our extensive ablation study provides further details on our design choices and also demonstrates a possibility of using a small, non-learnable, hand-crafted localization network that still out-competes many existing methods, while a learnable domain-agnostic network trained solely on synthetic data is able to improve upon this and achieve state-of-the-art results.

The remainder of this paper is structured as follows: in Section~\ref{sec:related_work} we present existing literature on counting and localization, in Section~\ref{sec:method} we detail our proposed approach using center-direction regression together with the localization network, while in Section~\ref{sec:experiments}, we demonstrate the performance of our method on several datasets and with a detailed ablation study. We conclude with a discussion in Section~\ref{sec:conclusion}.
\begin{figure*}
    \centering
    \includegraphics[width=\linewidth]{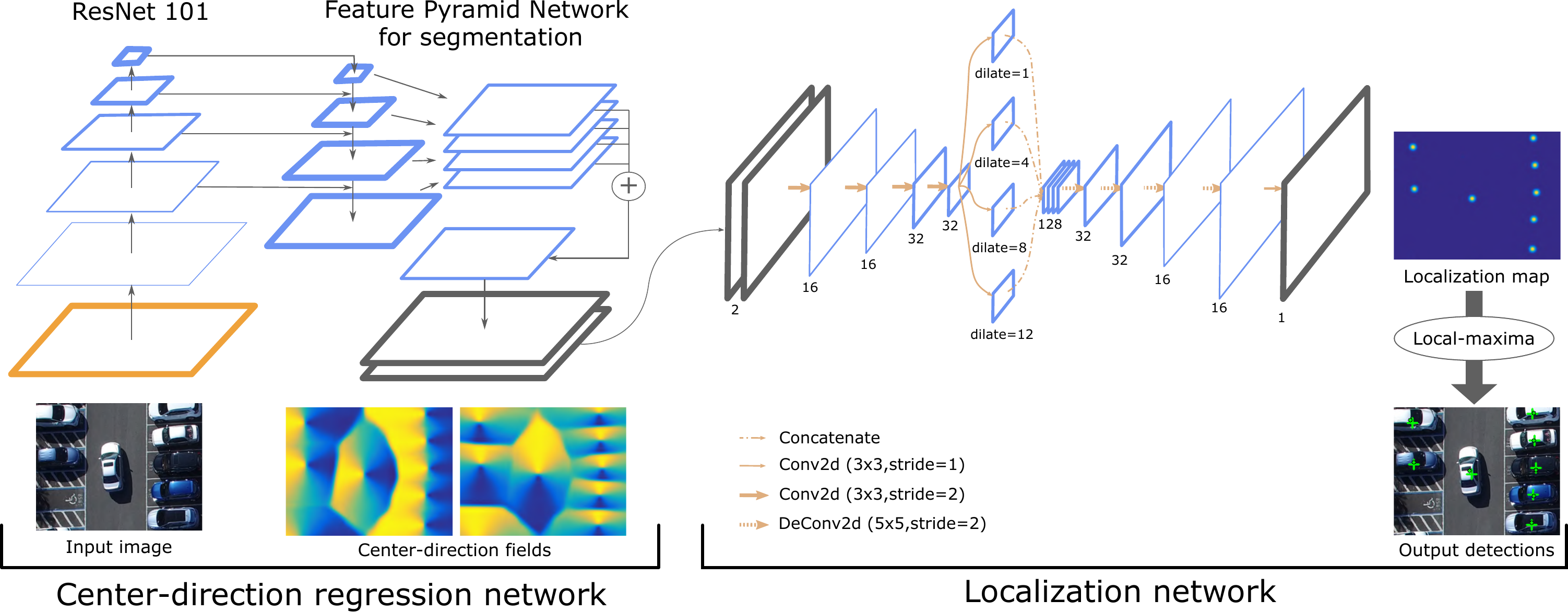}
    \caption{Architecture overview of dense center-directions regression and localization networks.} 
    \label{fig:arch}
\end{figure*}
\section{Related work}
\label{sec:related_work}

Object counting has been well-explored in the literature. While early approaches were based on hand-crafted features~\cite{Chan2008}, machine learning~\cite{Foroughi2015} and simple neural networks~\cite{Schofield1996}, more recent approaches rely heavily on deep learning. Modern approaches that rely on deep learning can be split into two groups: counting by regression and counting by detection. In the first group, objects are counted by directly or indirectly regressing the count number, while in the second group, the count number is obtained by detecting each individual object.

\paragraph{Counting by regression} Many modern state-of-the-art approaches to object counting employ regression of density estimation as an intermediate step to object counting. Instead of directly regressing the count number, they estimate the spatial probability map of objects and return the count number by integrating the estimated density~\cite{Zhang2016b}. In modern approaches, direct regression of the count number is occasionally used only as an auxiliary loss~\cite{Nguyen2022,Neven2019}. For training, only point supervision is required, but certain approaches~\cite{Cholakkal2019,Lei2021} also utilize only image-level annotations of the number of objects. Density estimation approaches are well suited for counting in densely crowded scenes~\cite{Wang2020,Wan2020}, particularly when combined with large dilation factors to increase the receptive field size~\cite{Li2018c}. However, sparse and background regions often present challenges for density estimation. Liu et al.~\cite{Liu2018} proposed to address these issues with the attention guided network that combines density estimation with counting-by-detection for regions with fewer objects. Rong et al.~\cite{Rong2021} also addressed problems of estimating density in background regions by proposing a background-aware loss that effectively adds a from-coarse-to-fine progressive attention mechanism. While improved performance in sparse regions has allowed density estimation methods to achieve state-of-the-art result for dense crowd counting~\cite{Wan2020,Rong2021}, such methods remain limited to applications where localization of the counted objects is not required. This is a direct consequence of relying on the density map, where the count number is encoded over a larger spread of the probability locations.

\paragraph{Counting by detection} On the other hand, approaches that employ counting-by-detection explicitly detect all objects being counted, and are thus capable of performing both localization and counting. Applying general object detectors with bounding box predictions can be a simple solution, but such detectors are sensitive to higher-density scenes due to overlapping bounding boxes~\cite{Goldman2019}. This can be improved by using Soft-IoU and Gaussian mixture model to reduce the overlap between the detected bounding boxes as shown by Goldman et al.~\cite{Goldman2019}. Hsieh et al.~\cite{Hsieh2017} also explored extending a region proposal-based method with a layout proposal network to capture the statistics in neighboring object locations for drone-based scenes.

To completely eliminate the problem of overlapping regions in dense scenes, recent methods directly predict object center positions instead of predicting bounding boxes. Such methods also significantly reduce the required labeling effort since only point annotations are needed. Wang et al.~\cite{Wang2021} predict the probability map of object center location from a Gaussian smoothed center points and relies only on point-supervision by using a self-training approach that estimates and infers pseudo-groundtruth sizes from point annotations. Ribera et al.~\cite{Ribera2019} proposed not to use object sizes during training at all and instead employed the weighted Hausdorff distance to learn an object location probability map. For object counting, their method further employed direct regression of the counting number from U-Net features merged with the predicted center probability map. Tong et al.~\cite{Tong2021} also used less labor-intensive point-supervision for predicting object locations while also estimating segmentation map for counting and segmentation of trees from drone-based imagery. The most recent work~\cite{DeArruda2022} employed a pyramid pooling model and multi-sigma refinement approach to achieve the state-of-the-art results for counting and localization from point supervision.

\paragraph{Center-point prediction methods}
Our proposed approach also employs point supervision and can be considered a counting-by-detection method, however, it differs from existing methods in the regression of the center locations. Instead of directly regressing the probability map of object center locations, as done in related methods~\cite{Ribera2019,Wang2021,Tong2021,DeArruda2022}, we propose to regress only the center direction and then use a lightweight domain-agnostic network to find the object centers. 

Regression of center locations has also been used in object detection~\cite{Zhou2019, Law2020, Duan2022, Wang2021}, segmentation~\cite{Neven2019} {and human pose estimation~\cite{Zhou2021}}. CenterNet~\cite{Zhou2019}, CenterNet++~\cite{Duan2022} and CornerNet~\cite{Law2020} perform regression of the object's center position as well as additional key-points to define the object bounding box. The same approach was also applied to 3D object detection from point clouds~\cite{Wang2021}. {While these methods were designed for other tasks, their first stage can still be used for center detection in point-supervision. CenterNet~\cite{Zhou2019} and the method proposed by Zhou et al.~\cite{Zhou2021} also perform direct regression of offsets to the center position, next to regression of binary probability map, however, they propose this only as a refinement of a location found initially through the center likelihood map. In CenterNet, offsets are trained only at locations where centers are positioned, while ignoring regression of offsets from surrounding locations as we do. Zhou et al.~\cite{Zhou2021}, as well as Neven et al.~\cite{Neven2019}, on the other hand perform prediction of vectors that point to the center from neighboring pixels, similar to our approach. However in \cite{Neven2019}, center positions are not learned directly but through an IoU-loss and only within the instance mask, making this method incompatible with point-supervision training as it requires pixel-wise segmentation labels for the ground truth. Methods by Zhou et al.~\cite{Zhou2021} and Neven et al.~\cite{Neven2019} also do not formulate the center direction as we do to enable localization with lightweight network, but instead require complex post-processing with clustering~\cite{Neven2019} or Hough voting~\cite{Zhou2021}, making them unsuitable for localization of large number of objects common in counting.}

\section{Object counting and localization with center-directions}
\label{sec:method}

We propose a novel method for object counting and localization using a dense prediction of directions to object centers. Instead of directly predicting a binary probability of a center for each location, as is commonly performed in other approaches~\cite{Wang2021}, we propose to first predict center-directions and then use domain-agnostic localization network to count objects and find their position (c.f. Fig.~\ref{fig:arch}). We term the proposed approach with Center-Direction Regression as \emph{CeDiRNet}.

\subsection{Dense regression of center-directions}

\paragraph{Parametrization} We propose to first predict a direction to the closest center position for each object in an image from all of its surrounding pixel locations, i.e., for each pixel location in an image, we predict a direction that points towards the closest object center. For each pixel location $\mathbf{x}_{i,j}=(i,j)$, we can define $\phi_{i,j}$ as a direction angle pointing towards the closest object center $\mathbf{y}=(n,m)$:

\begin{align}
    \phi_{i,j} &= \tan \bigg( \frac{m-j}{n-i} \bigg).
\end{align}

\begin{figure}
    \centering
    \includegraphics[width=\linewidth]{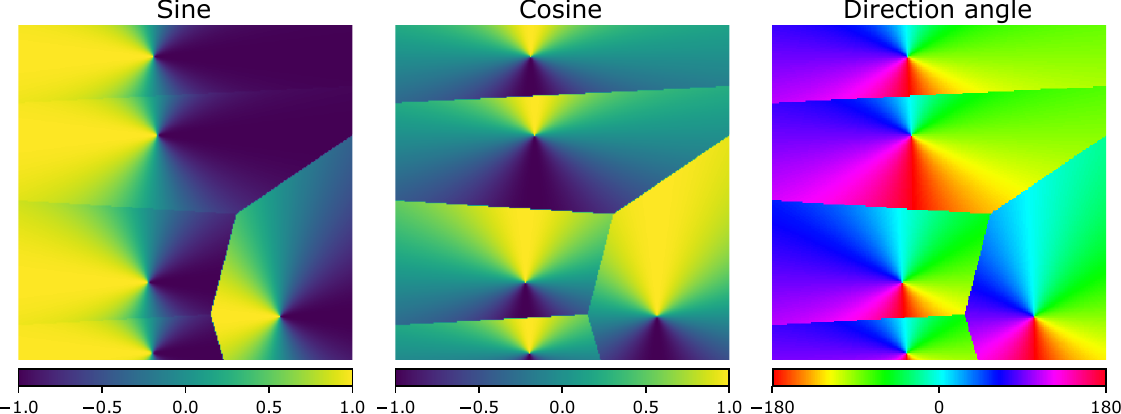}
    \caption{Example of center-directions fields in re-parametrized form: $C_{sin}$ (left) and $C_{cos}$ (middle). Last column is a visualization of the actual angle $\phi$ obtained by $atan2(\cdot)$}
    \label{fig:center_vector_vis}
\end{figure}

We propose a re-parameterization of the direction angle $\phi$ using $sin(\cdot)$ and $cos(\cdot)$, and we term $\phi$ in this form as \emph{center-direction}. Direction angle and their corresponding re-parameterized fields are depicted in Fig.~\ref{fig:center_vector_vis}.  A deep learning model for dense prediction is then used to perform a dense regression of center-direction for every pixel location in the image. Our proposed model can therefore be described as regression of two dense fields $\hat{C}_{sin},\hat{C}_{cos}\in\mathbb{R}^{N\times M}$ from an input image $\mathcal{I}\in\mathbb{R}^{N\times M}$, corresponding to $sin(\phi)$ and $cos(\phi)$ of the closest center point: 
\begin{align}
    \hat{C}_{sin} &= ConvNet_{sin}(\mathcal{I}), & 
    \hat{C}_{cos} &= ConvNet_{cos}(\mathcal{I}).
\end{align}

A re-parametrized center-direction affords multiple benefits. It eliminates the issue of regressing cyclical values in angle while also translates values into bounded range of $[-1,+1]$. This is more suitable for direct regression than using unbounded vectors in Cartesian space~\cite{Neven2019}, since all values will be of the same orders of magnitude, thus having similar gradients. This also eliminates the use of any additional scaling factor that is often used to mitigate the issue of unbounded range in Cartesian space.

We can also re-obtain the original angle $\phi$ from the regressed $\hat{C}_{sin}$ and $\hat{C}_{cos}$ by applying inverse tangent function. For mathematical stability, we use $atan2(\cdot)$ function, however, to find the final object location we do not need angle directly and use it only for visualization purposes.

\paragraph{Architecture}

To learn center-direction regression for every pixel location, we use a deep network for dense prediction. In general, any network for dense prediction would suffice, such as hourglass architectures or encoder-decoder models. In this paper, we use an encoder-decoder model with ResNet-101 as encoder and Feature Pyramid Network~\cite{Lin2016b,Li2019c} as decoder. The architecture is implemented in PyTorch based on~\cite{Yakubovskiy:2019}, where FPN consists of four levels of the feature pyramid with an additional segmentation head consisting of four layers that perform the final up-sampling and regression\footnote{\url{https://github.com/vicoslab/CeDiRNet.git}}.

\paragraph{Learning}

The regression model for center-directions, i.e.,  $ConvNet_{cos}$ and $ConvNet_{sin}$, is learned in a fully supervised manner. For each regressed location $\mathbf{x}_{i,j}$, we define a ground truth regression target as a center-direction to the closest object in a re-parametrized form, $C_{sin}(\mathbf{x}_{i,j})$ and $C_{cos}(\mathbf{x}_{i,j})$. The final loss function is formulated using the per-pixel $L_1$ distance:
\begin{align}
\label{eq:polar_loss}
    \mathcal{L} =& \sum_{i,j}W(\mathbf{x}_{i,j}) \cdot |\hat{C}_{sin}(\mathbf{x}_{i,j}) - C_{sin}(\mathbf{x}_{i,j})| + \\ 
                 & \sum_{i,j}W(\mathbf{x}_{i,j}) \cdot |\hat{C}_{cos}(\mathbf{x}_{i,j}) - C_{cos}(\mathbf{x}_{i,j})|,
\end{align}
where $W(\mathbf{x}_{i,j})$ defines a per-pixel weight assigned for each location $\mathbf{x}_{i,j}$, which also acts as an averaging term, and $| \cdot |$ represents a per-pixel absolute error. We use the per-pixel weight to balance the losses between the pixels closer to the object and ones farther away. Since the number of pixels farther from the center will always be much larger than the number of pixels around the center, this will result in a larger influence of distant pixels, thus reducing the speed of learning for pixels around the center. However, during learning more emphasis should be given to proper regression of center-directions around the object center. We re-balance this by assigning larger weights to the pixels around the object centers: 
\begin{align}
\label{eq:weigth_normalization}
    W(\mathbf{x}_{i,j}) = 
    \begin{cases}
        \dfrac{1}{K \cdot | \mathcal{F}_k | }  & \text{if } \mathbf{x}_{i,j}\in\mathcal{F}_k \\ 
        \dfrac{1}{|\mathcal{B} | } & \text{if } \mathbf{x}_{i,j}\in\mathcal{B},\\
    \end{cases}
\end{align}
where $\mathcal{B}$ represents the set of distant pixels, $\mathcal{F}_k$ represents the set of pixels around the k-th object center and $K$ is the total number of objects in an image. Division between pixels belonging to $\mathcal{F}_k$ and $\mathcal{B}$ is completely independent from the actual objects' visual boundaries since we only require loosely defined areas of higher importance. We define $\mathcal{F}_k$ and $\mathcal{B}$ based on the distance to the closest center $\mathbf{y}_k$ using a cutoff threshold $\epsilon$:
\begin{align}
\label{eq:weigth_normalization}
    \mathcal{F}_k = & \{\mathbf{x}_{i,j} \mid (\mathbf{x}_{i,j} - \mathbf{y}_k)^2  < \epsilon\}, \\
    \mathcal{B} = & \{\mathbf{x}_{i,j} \mid \forall k \in \{1..K\}, (\mathbf{x}_{i,j} - \mathbf{y}_k)^2 \ge \epsilon \}.
\end{align}
The cutoff distance threshold $\epsilon$ could be set relative to the image size and object density, but in our experiments we used a single fixed $\epsilon$, which has proven sufficient for different datasets. We provide more details on different $\epsilon$ values in the ablation study.

The proposed weighing scheme effectively normalizes the losses between the pixels around the object centers,  $\mathcal{F}_1 \cup \mathcal{F}_2 \cup .. \cup \mathcal{F}_K$, and more distant ones, $\mathcal{B}$, as well as between the individual sets $\mathcal{F}_k$. This is implemented as normalization with the number of pixels in $\mathcal{B}$ set for $\mathbf{x}_{i,j}\in\mathcal{B}$, and as normalization with the number of pixels in each $\mathcal{F}_k$, averaged over the number of objects $K$, for $\mathbf{x}_{i,j}\in \mathcal{F}_1 \cup \mathcal{F}_2 \cup .. \cup \mathcal{F}_K$.

\subsection{Object localization from center-directions}

Although a single regressed center-direction value does not have enough information to exactly locate the object's position, the information about the exact center location arises when center-directions from surrounding pixels are considered. Theoretically, the object center can be found as an intersection point between multiple lines obtained from the regressed direction angles from multiple surrounding pixel locations. However, directly applying an analytical solution to find intersection points from a large set of lines (i.e., a large image will contain hundreds to millions of lines) would not be efficient nor robust, particularly when a large number of objects is present in the image.

We instead propose to find object centers by applying an additional small network directly to the re-parameterized center-direction fields $\hat{C}_{sin},\hat{C}_{cos}$, and output a dense probability map of object centers $\mathcal{\hat{O}}\in\mathbb{R}^{N\times M}$:
\begin{align}
    \hat{\mathcal{O}}_{cent} &= ConvNet_{cent}(\hat{C}_{sin},\hat{C}_{cos}).
\end{align}

Since $C_{sin}$ and $C_{cos}$ result in a heatmap image with extremely unique visual signature around each object center (c.f. Fig.~\ref{fig:center_vector_vis}), we can construct a network for $ConvNet_{cent}$ that outputs a high probability at the exact center position and a low probability everywhere else. Finding local-maxima values in the resulting $\hat{\mathcal{O}}_{cent}$ yields the number of objects in an image and their center positions, which is then used as the final output of our model:
\begin{align}
     positions(\mathcal{I}) &= \argmax_{i,j}\left[ \text{LocalMax}(\hat{\mathcal{O}}_{cent})\right],\\
     count(\mathcal{I}) &= \sum_{i,j}\left[\text{LocalMax}(\hat{\mathcal{O}}_{cent})\right].
\end{align}

A simple local maxima detection is sufficient to extract the locations from the regressed output heatmap. This stems directly from applying the localization network to center-directions fields with the limited range of values and a unique visual signature around each object center thus resulting in the localization output $\hat{\mathcal{O}}_{cent}$ with minimal noise even for locations where regressed center-directions are somewhat noisy. This removes any need for an additional complex post-processing heuristics with thresholding, smoothing or clustering.

\paragraph{Architecture}

For the center localization network, we use an hourglass architecture consisting of four levels in both encoder and decoder, i.e., eight levels in total. Each level consists of a single convolution block (Conv2D with batch normalization and ReLu) with 16 or 32 channels. To increase the receptive field size, we add a middle layer between the encoder and decoder containing multiple parallel convolutions, each with a different dilation factor similar to ASPP module in DeepLab~\cite{Chen2016a}. We use the dilation factors of $1$, $4$, $8$ and $12$. The architecture is depicted in Fig.~\ref{fig:arch}.

\paragraph{Learning}

The localization network is learned independently of the center-direction regression. Localization network is learned with a per-pixel loss function. We define a per-pixel ground truth regression target $\mathcal{O}_{cent}(\mathbf{x}_{i,j})$ as a Gaussian distance to the closest object center point $\mathbf{y}$:
\begin{align}
\label{eq:centernet_gt}
    \mathcal{O}_{cent} (\mathbf{x}_{i,j}) = min \Bigg( 1.0,\exp{ \Bigg(-\cfrac{|\mathbf{x}_{i,j}-\mathbf{y}|-\xi}{2\sigma^2} \Bigg)} \Bigg),
\end{align}
where we set $\sigma=2.5$ and $\xi=1$ to ensure more than one pixel in the center have maximum value and to additionally increase the extent of the Gaussian. The network parameters $ConvNet_{cent}$ are then optimized using $L_1$ loss:

\begin{align}
\label{eq:centernet_loss}
    \mathcal{L}_{cent} =& \dfrac{1}{N \cdot M} \sum_{i,j}W_{cent}(\mathbf{x}_{i,j}) \cdot | \hat{\mathcal{O}}_{cent}(\mathbf{x}_{i,j}) - \mathcal{O}_{cent}(\mathbf{x}_{i,j}) |, 
\end{align}
where $N \cdot M$ is the number of pixels in an image, $W_{cent}(\mathbf{x}_{i,j})$ is an additional per-pixel weight and $| \cdot |$ is a per-pixel absolute error. Since the number of background pixels is significantly larger than the number of foreground pixels, we increase the weight for non-background pixels using $w_{fg}>1$:
\begin{align}
\label{eq:centernet_weigth_normalization}
    W_{cent}(\mathbf{x}_{i,j}) = 
    \begin{cases}
        w_{fg} & \text{if } \mathcal{O}_{cent}(\mathbf{x}_{i,j}) > 0 \\
        1 & \text{otherwise}.\\
    \end{cases}
\end{align}

\paragraph{Domain-agnostic learning using synthetic data}
\textcolor{blue}{
The localization network is learned in a fully domain-agnostic manner without the need for real datasets. Our approach translates domain-specific images into a domain-agnostic center-directions map ($C_{sin}$ and $C_{cos}$), effectively separating the localization task from domain-specific features. This allows independent training across domains using synthetic data, simplifying the overall learning procedure as the localization network no longer relies on domain-specific datasets. Simultaneously, leveraging synthetic data mitigates overfitting issues by diversifying the training dataset, as it encompasses random generation of the number and position of objects in the input data. This approach prevents overfitting associated with a fixed number of objects and locations in real data, leading to enhanced and more robust performance. Additionally, we introduce Gaussian noise and occlusions to the synthetic data to enhance robustness. Our ablation study demonstrates that this approach yields better performance compared to finetuning on a specific domain with real data thus completely eliminating the need for real data.}

\section{Experiments}
\label{sec:experiments}

In this section, we report results of an extensive evaluation of the proposed method. Evaluation is performed on six datasets for object counting in remote sensing domains: Sorghum plant~\cite{Ribera2019}, CARPK~\cite{Hsieh2017}, PUCPR+~\cite{Hsieh2017}, Acacia-6~\cite{Tong2021}, Acacia-12~\cite{Tong2021} and Oil palm~\cite{Tong2021}. We first report results obtained on all these datasets and compare the proposed method with the related work. This is followed by an ablation study, analyzing the influence of different hyperparameters and design choices we made.

\subsection{Datasets}

\paragraph{Sorghum plant dataset} Sorghum plant dataset provided by Ribera et al.~\cite{Ribera2019} consists of 60,000 images of sorghum plants captured from a top-down perspective from a UAV and point-annotation of each plant. The dataset contains images of a single physical sorghum field that were first stitched together into a single $6000 \times 12000$ orthoimage and then split into separate training, validation and testing regions. Final images are then randomly cropped patches from each region, resulting in 50,000 training, 5,000 validation and 5,000 testing images. The height and width of the random crops are uniformly distributed between 100 and 600 pixels. Some of these image crops may overlap, but they are restricted to their area. Image patches resized to $256\times 256$ pixels are also provided, however, we use only originally cropped images with preserved aspect ratios in our experiments, and additionally crop/pad them to $256 \times 768$ pixels for training without changing the aspect ratios. Some examples of images are shown in Figs.~\ref{fig:sorghum_detections} and~\ref{fig:sorghum_fp_detections}.

\paragraph{CARPK and PUCPR+ datasets} Car Parking Lot (CARPK) and the Pontifical Catholic University of Parana+ (PUCPR+) are car-counting datasets that contain images of parking lots provided by Hsieh et al.~\cite{Hsieh2017}. CARPK dataset consists of 1,444 high-resolution images containing 89,777 annotated cars that are split into 989 training and 495 testing images. PUCPR+ dataset consists of 125 high-resolution images containing 16,456 annotated cars that are split into 100 training and 25 testing images. All images in CARPK dataset are captured from the top-down perspective taken with an UAV drone, while images in PUCPR+ dataset are taken by a fixed camera mounted in front of a parking lot. Some examples of images are shown in Figs.~\ref{fig:carpk_detections} and~\ref{fig:sorghum_fp_detections}. In both datasets objects are annotated with bounding boxes, however, in our experiments we use only the center of the bounding box and discard object sizes. Since a validation set is not provided, we selected a subset of training images for validation purposes. For PUCPR+, we randomly selected 10 images, while for CARPK, we selected a set of all training images that originated from the same video sequence to ensure minimal intersection between the training and the validation sets. This resulted in using 155 validation images for CARPK dataset. Nevertheless, the visual appearance of the validation set is much closer to the appearance of the training set than the testing set, making the validation set less than ideal. 

\paragraph{Acacia and Oil palm datasets} Acacia and Oil palm datasets consists of three datasets provided by Tong et al.~\cite{Tong2021}: Acacia-6, Acacia-12 and Oil palm. Each dataset consist of a large orthoimage of tree plantations. Images are of $31,644 \times 26,420$ pixels in size for Acacia-6, $36,521 \times 31,181$ pixels for Acacia-12 and $64,273\times 27,839$ pixels for Oil palm. The three datasets contain 18,528, 25,691 and 130,035 annotated objects for Acacia-6, Acacia-12 and Oil palm, respectively, and each object is annotated with a single center point only. Since authors do not provide cropped images nor the train/test splits that were used in their experiments, we first extracted sliding window patches and then split them into three non-overlapping sets for 3-fold cross validation. Sliding window patches of $512 \times 512$ pixels in size were extracted with a step of 384 pixels, while ignoring patches that do not contain any valid image pixels. This resulted in 1779, 2495 and 11,867 images for Acacia-6, Acacia-12 and Oil palm, respectively. Some examples of image patches are shown in Fig.~\ref{fig:plantation_trees_detections}. We then assigned the cropped images into three non-overlapping sets based on the original large image that is split into three regions. For each fold in the cross-validation that corresponds to one region, the testing images corresponded to all patches that contain only pixels from that region, while the remaining patches that do not contain any pixels from that region were used for training and validation. For each dataset, the original large image was split into three regions either in vertical or horizontal orientation to ensure similar image count for each fold. We perform evaluation on this dataset by merging the results from each test patch from all folds into the original image. 

\subsection{Evaluation metrics}

 We use standard metrics for evaluating object counting methods. This includes the mean absolute error (MAE) and root mean squared error (RMSE):
\begin{align}
\label{eq:mae_and_rmse}
    MAE &= \frac{1}{N}\sum_{i=1}^N \big|c_i - \hat{c}_i\big|, &
    RMSE &= \sqrt{\frac{1}{N}\sum_{i=1}^N \big( c_i - \hat{c}_i \big)^2},
\end{align}
where $c_i$ is the actual number of objects in the $i$-th image and $\hat{c}_i$ is the predicted number of objects. We additionally evaluate localization performance since the proposed method returns not only the number of objects but also the exact object location. Localization metrics are often stricter than MAE or RMSE, since a low count error can be achieved even with a large number of erroneous detections if the number of false positives and false negatives balance out. As a localization metric, we thus measure $F_1$, precision and recall averaged over all images:
\begin{align}
\label{eq:F1_ap_ar}
    F_1 &= \frac{1}{N}\sum_{i=1}^N \frac{2\mathit{TP}_i}{2\mathit{TP}_i+\mathit{FP}_i+\mathit{FN}_i},
\end{align}
\begin{align}
\label{eq:ap_ar}
    precision &= \frac{1}{N}\sum_{i=1}^N \frac{\mathit{TP}_i}{\mathit{TP}_i+\mathit{FP}_i},&
    recall &= \frac{1}{N}\sum_{i=1}^N \frac{TP_i}{TP_i+FN_i},
\end{align}
where $\mathit{TP}_i$, $\mathit{TN}_i$, $\mathit{FP}_i$ and ${FN}_i$ are calculated for the $i$-th image at a specific score threshold, which is obtained as the threshold achieving the best $F_1$ and MAE score on the validation set. True detections are considered only when the distance to the ground truth location is less than $\tau$ and is calculated as a linear sum assignment problem~\cite{Kuhn1955} between the ground truth locations and the predicted object locations. Since different related methods use different $\tau$ thresholds, we report metrics at multiple $\tau$ for fair comparison. However, we observed that some methods report values at small $\tau=5$, which considering images of around $512\times 512$ is unrealistically small margin of error. We find the metric at around $20\leq\tau\leq40$ to be a more realistic reflection of true performance, particularly in high-resolution images. 

\subsection{Implementation details}

\paragraph{Learning}

Both network architectures were trained with Adam optimizer using the learning rate of $10^{-4}$. Additionally, the learning rate decay was applied at the start of each epoch with a factor of $(1-n)^{0.9}$, where $n$ is the fraction of completed epochs. For the center-direction regression network, we used a batch size of 32 and trained the network for 200 epochs, except for the Sorghum dataset, where we used a batch size of 128 and trained it for 50 epochs due to smaller images but larger dataset. We used ResNet backbone pre-trained on ImageNet dataset as provided by PyTorch framework, while FPN layers were initialized to the normal distribution based on Glorot and Bengio~\cite{Glorot2010} method with a gain value of $0.1$. For all datasets, training was distributed over 8 GPUs. For the center localization network, we used a batch size of 768 and trained the network for 200 epochs on 16 GPUs. We set $w_{fg}=50$, which we found to perform the best. We used 5000 synthetically generated images in each epoch. 

\paragraph{Data augmentation} For all datasets, we apply data augmentation using random image rotation and color pixel jittering. Rotational augmentation is implemented as a vertical and a horizontal image flip, each being performed randomly with a probability of $0.5$. We then apply random color jittering with a probability of 0.5, where added brightness, color and saturation jitter are chosen uniformly from $[0.8, 1.2]$, while hue jitter is chosen uniformly from $[-0.2, 0.2]$.

\begin{table}[]
    \centering
    \begin{adjustbox}{width=1\linewidth,center} 
        \begin{tabular}{c|cccc|cc}
            \toprule
            \textit{Method}     & $\tau$ [px] &  prec [\%] & recall [\%] & $F_1$ [\%] & MAE & RMSE \\
            \midrule
            \midrule
            Tong et al.~\cite{Tong2021} &  - & 92.60 & 93.00 & 92.60 & 2.33 & 3.43  \\            
            \midrule
            {CenterNet~\cite{Zhou2019}}  & {40} & {95.18} & {96.22} &	{95.58} & \multirow{4}{*}{{2.16}} & \multirow{4}{*}{{3.41}} \\
            {CenterNet~\cite{Zhou2019}}  & {30} & {95.11} & {96.16} &	{95.52} &  &  \\
            {CenterNet~\cite{Zhou2019}}  & {20} & {95.06} & {96.10} &	{95.47} &  &	\\
            {CenterNet~\cite{Zhou2019}}  & {15} & {95.02} & {96.06} & {95.43} \\
            \midrule
            {CenterNet~\cite{Zhou2019}} & {5} & {89.53} & {90.50} & {89.91} & {2.16} & {3.41} \\
            Faster R-CNN~\cite{Ribera2019} & 5 & 86.60 & 78.30 & 82.20 & 9.40 & 17.70 \\
            Ribera et al.~\cite{Ribera2019} & 5 & 88.10 & 89.20 & 88.60 & 1.90 & \textbf{2.70} \\ 
            CeDiRNet (our)  & 5 & \textbf{89.76} & \textbf{91.03} & \textbf{90.32} & \textbf{1.78} & 2.99 \\
            \midrule
            CeDiRNet (our)  & 15 & 95.80 & 97.17 & 96.41 &
            \multirow{4}{*}{1.78} & \multirow{4}{*}{2.99} \\
            CeDiRNet (our)  & 20 &	95.83 &	97.20 &	96.44 &  &	\\
            CeDiRNet (our)  & 30 & 95.86 & 97.24 &	96.47 &  &  \\
            CeDiRNet (our)  & 40 & 95.91 & 97.29 &	96.52 &  &  \\
            \bottomrule
        \end{tabular}
    \end{adjustbox}
    \caption{Results on Sorghum dataset with different $\tau$ values. Note that~\cite{Tong2021} do not explicitly report which $\tau$ value was used for Sorghum dataset but they use $\tau=40$ for other datasets.}
    \label{tab:sorghum}
\end{table}

\paragraph{Synthetic data for learning localization network}

For the localization network, we used synthetic training images that were randomly generated during each iteration. For each sample, we randomly assigned a number of center points and generated $C_{sin},C_{cos}$ as input images, each $512\times512$ pixels in size. The number of points were randomly selected from the uniform distribution of $\mathcal{U}(5,50)$. We also added additional center points around the originally generated ones. Around each existing point, we added additional points with a probability of $0.25$, while the number of added points around each one was randomly sampled from the uniform distribution of $\mathcal{U}(2,5)$. We additionally simulated degradation of input data to minimize the problem of overfitting to the synthetic appearance. We added zero-mean Gaussian noise with the probability of $0.25$, where the standard deviation of the noise was randomly selected from the uniform probability of $\mathcal{U}(0.1,2)$. Noise was additionally smoothed with a Gaussian blur using $\sigma=3$. Finally, we simulated occlusion of the center-directions around the center point. Occlusions were added to each center point independently with the probability of 0.75. We used circular occlusions positioned around the existing center points. Positions were also randomly perturbed with the probability of 0.5. The size of the circular occlusions was randomly selected in a range of $[5 ,0.025\cdot D]$ pixels, where $D=\sqrt{W^2+H^2}$ is the diagonal of the generated image. The selection of sizes was also slightly skewed towards larger images.

\begin{figure*}
    \centering
    \includegraphics[width=\linewidth]{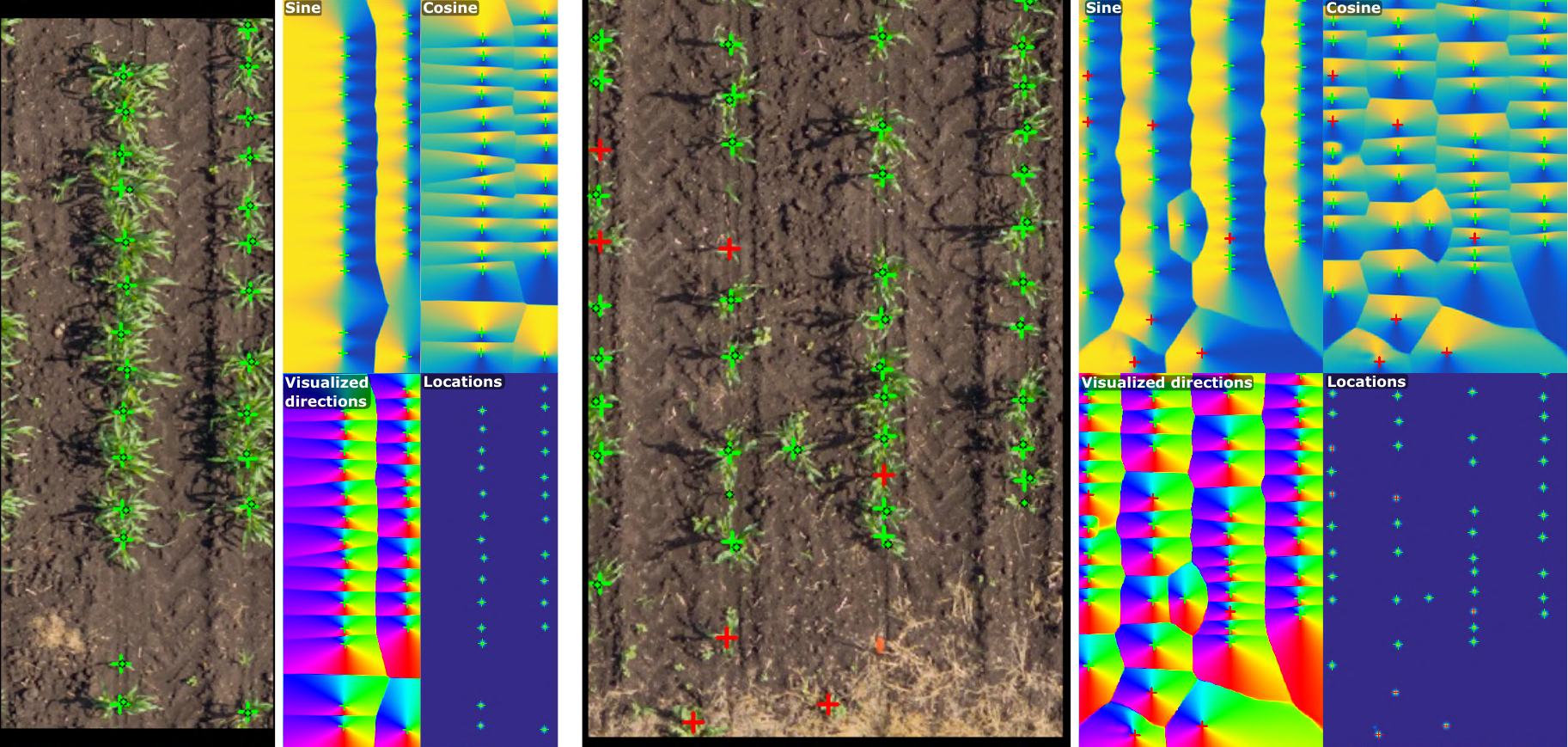}
    \caption{Examples of regressed center-directions and localization detections on Sorghum dataset. Green and red crosses are correct and incorrect detections, respectively, shown based on $\tau=40$, while ground truths are shown with green dots.}
    \label{fig:sorghum_detections}
\end{figure*}
\paragraph{Hard samples mining within epoch}

We additionally enhanced the learning process with a mechanism that gives more emphasis on the learning of hard samples. We implemented this by augmenting the training batch with samples that are selected from a set of hard examples. At the start of each epoch, at minimum 10\% of the most difficult samples are selected as hard samples and during each iteration N$_{hr}$\% of the batch size is randomly selected only from those hard samples. The degree of difficulty of each sample is measured by its loss value as well as by using the actual number of false positives and false negatives, while giving slightly more emphasis on false negatives..

\paragraph{Hyperparameter optimization on validation}

For each dataset, we selected the model at the best performing epoch and score threshold, while we also optimized free hyperparameters. Both optimizations were performed on the \emph{validation set}. Two hyperparameters were optimized: (a) learning weight decay and (b) percent of batch size used for hard samples ($N_{hr}$). A hyperparameter search was performed for each dataset independently, and a model with the specific combination of parameters that performed the best was then used for evaluation on the testing set. For the weight decay, we explored using values $0$ and $10^{-4}$, while for hard samples mining, we explored using values $N_{hr}=0$ and $N_{hr}=50$\%. 

\subsection{Results}

Evaluation results are reported in Table~\ref{tab:sorghum} for Sorghum plant dataset, in Table~\ref{tab:carpk_and_pucpr} for CARPK and PUCPR+ datasets, and in Table~\ref{tab:plantation_trees} for Acacia-06, Acacia-12 and Oil palm datasets. The proposed model is shown to outperform all existing state-of-the-art models. All results reported in those tables were achieved with a generic localization network that was trained only on synthetic data. Domain-specific data were used only to train the center-direction regression network.

\begin{figure}
    \centering
    \includegraphics[width=.49\linewidth]{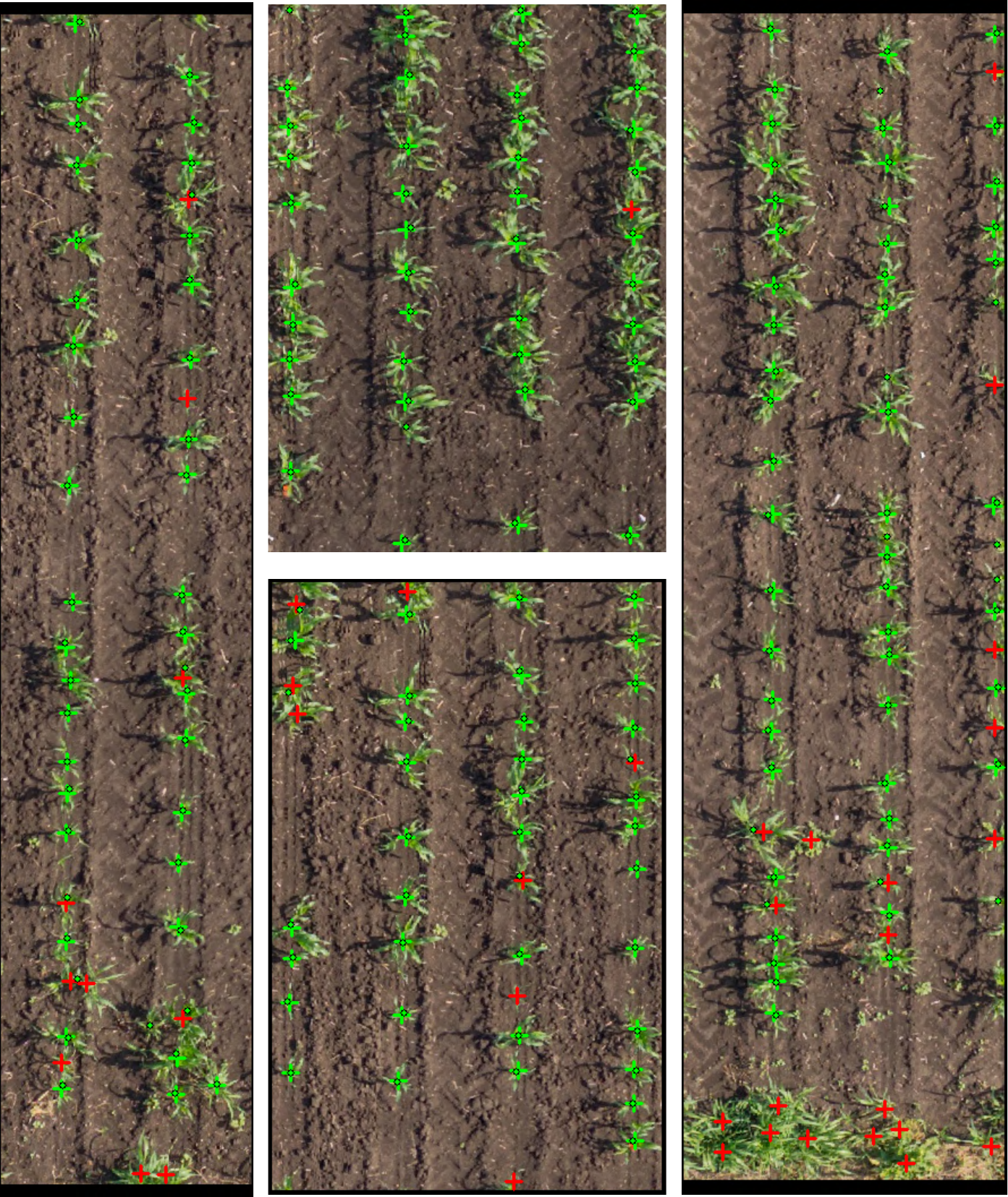}~\includegraphics[width=0.50\linewidth]{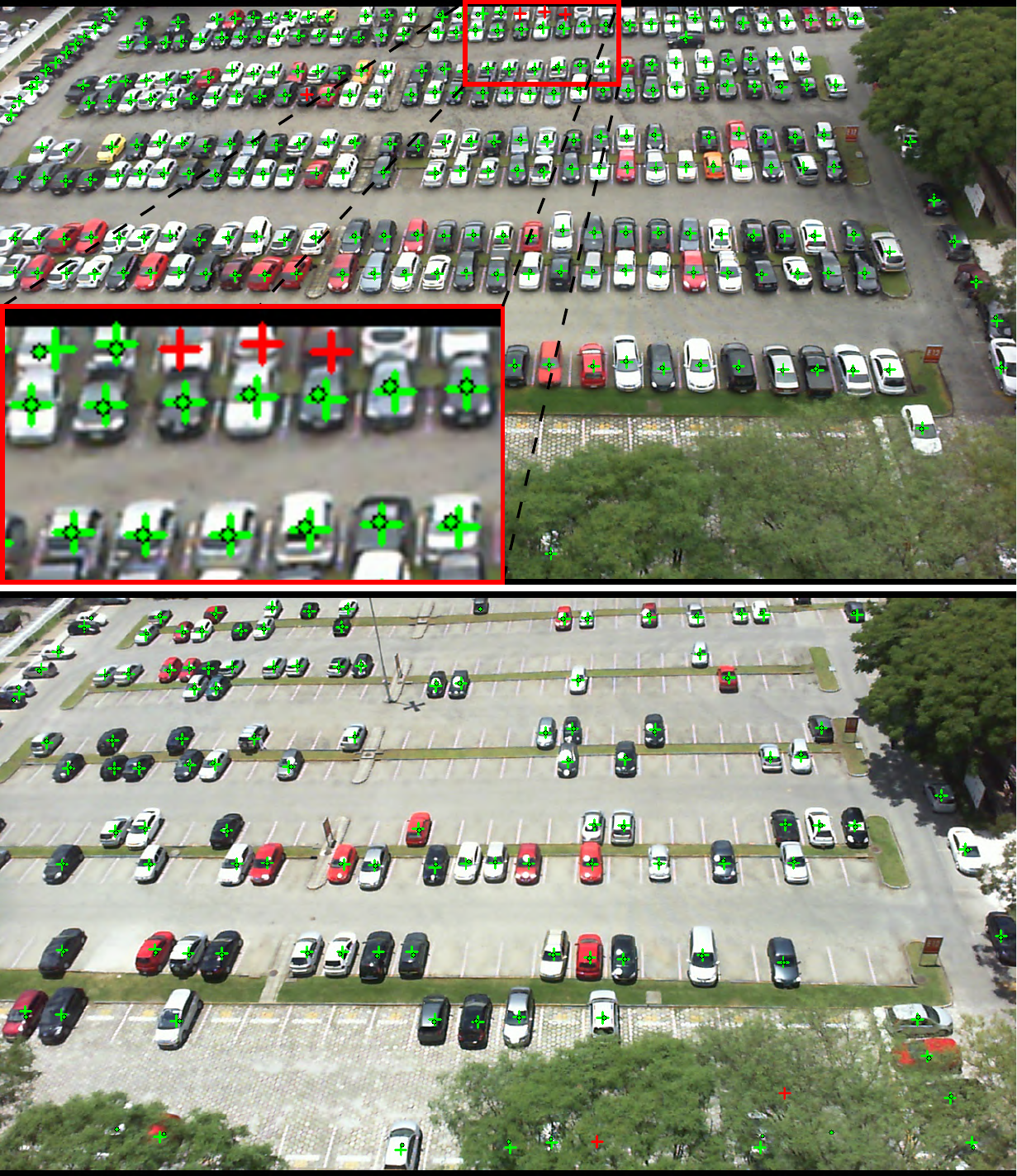}
    \caption{Examples of true and false detections on Sorghum dataset using $\tau=5$, on the left, and detections on PUCPR+ dataset using $\tau=40$, on the right. }
    \label{fig:sorghum_fp_detections}
\end{figure}

\begin{figure*}
    \centering
    \includegraphics[width=\linewidth]{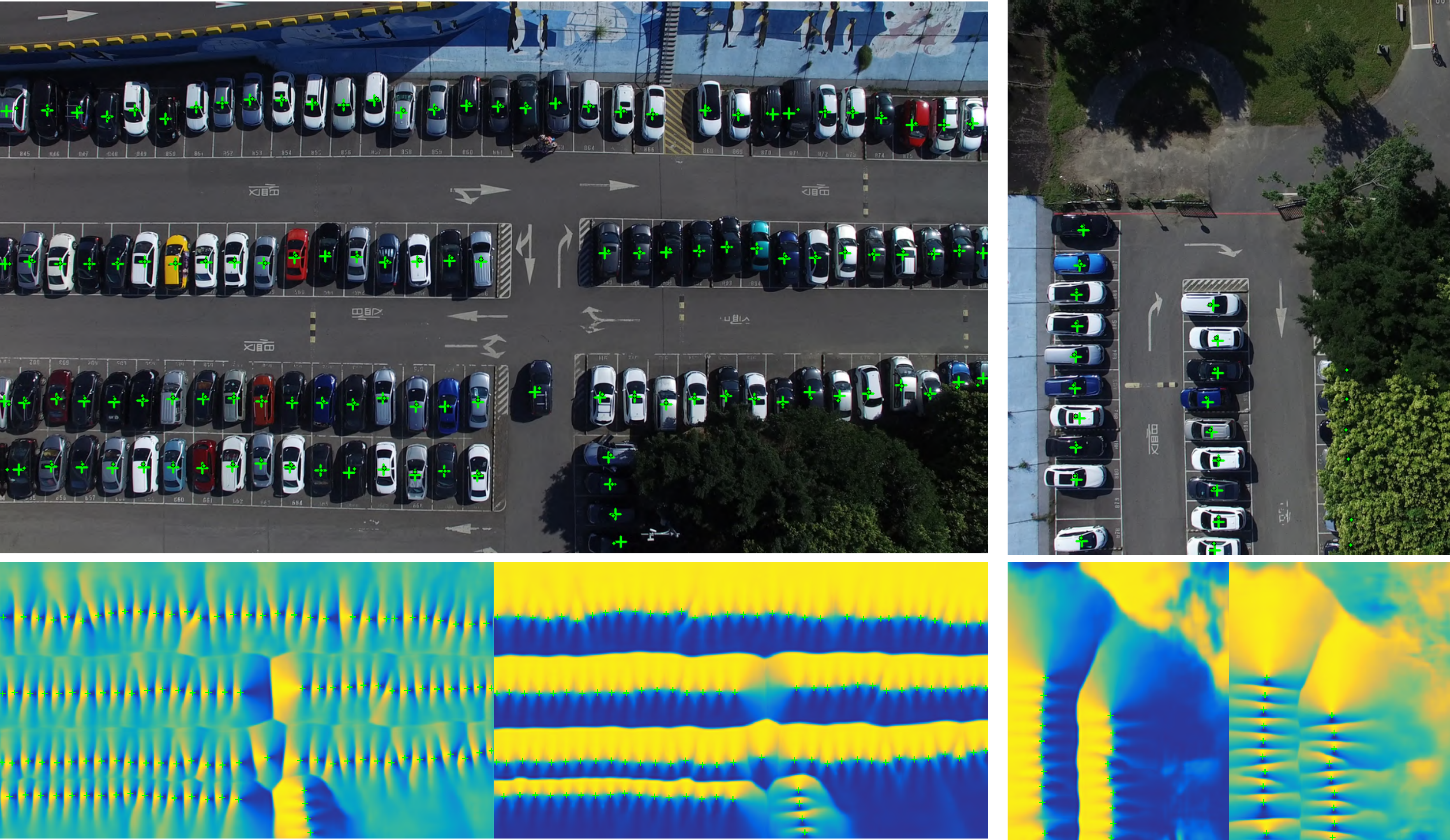}
    \caption{Examples of detections (top) and regressed center-directions (bottom) on CARPK dataset based on $\tau=40$}
    \label{fig:carpk_detections}
\end{figure*}
{\paragraph{Related methods} We compare the proposed method with a number of related approaches that are currently the best performing methods for each dataset as reported in the literature. We also include CenterNet~\cite{Zhou2019} as a closely related approach that combines two most common approaches for counting: (a) direct prediction of center locations as a heatmap trained using the binary cross-entropy loss, and (b) direct regression of offsets to the centers. Although CenterNet was designed for bounding box detection, we use only the first stage for center detection, which can be trained on point annotations only. For fair comparison, we use the same backbone and augmentation as for CeDiRNet (i.e., ResNet-101, color jittering and image mirroring), while following the same evaluation protocol. We also perform hyper-parameter optimization for the learning rate ($1.25\cdot10^4$ and $5\cdot10^4$), learning decay (no decay, or step-down at 65\% and 85\%), the best epoch and the threshold. Results are reported with hyper-parameters selected on the validation set.}

\paragraph{Sorghum} On Sorghum dataset the proposed model outperforms existing state-of-the-art point supervision approaches, both in terms of counting and in localization, as shown in Table~\ref{tab:sorghum}. In terms of counting, the CeDiRNet outperforms SOTA with MAE of 1.78 versus MAE of 1.90 for the second-best model (lower is better){, while CenterNet laggs behind with MAE of 2.16}. In RMSE metric, CeDiRNet slightly lags behind the SOTA model, but still outperforms Faster R-CNN, the method by Tong et al.~\cite{Tong2021} {and CenterNet}. Since RMSE penalizes larger errors more, this may be an indication that the proposed model has a larger variance of errors, i.e., fewer images with count errors, but each error is slightly larger. CeDiRNet outperforms all SOTA models when considering a less biased localization metric, where the center-direction model is able to achieve $F_1$ of 90.32\% while Ribera et al.~\cite{Ribera2019} achieves $F_1$ of 88.10\%. {CeDiRNet also outperforms CenterNet, with CenterNet lagging slightly behind at $F_1$ of 89.92\%.} Localization errors are reported at $\tau=5$, but we also report values at larger $\tau$ for easier comparison with other models. Tong et al.~\cite{Tong2021} did not explicitly report which $\tau$ was used for their Sorghum experiment, but they report using $\tau$ between 20 and 40 in their experiments. CeDiRNet achieves better localization than reported results already at $\tau\ge15$. {At a larger $\tau$, CeDiRNet also outperforms CenterNet by around 1 percentage point in all cases.}

\begin{table}[]
    \centering
    \begin{adjustbox}{width=1\linewidth,center}
        \begin{tabular}{cccccc}
            \toprule
              \multirow{2}{*}{\textit{Method}}   & \multicolumn{2}{c}{CARPK} & & \multicolumn{2}{c}{PUCPR+} \\
                 &  MAE & RMSE & & MAE & RMSE \\
            \midrule
            \midrule
            RetinaNet~\cite{Hsieh2017} & 16.62 & 22.3 & & 24.58 & 33.12 \\
            Goldman et al.~\cite{Goldman2019} & 6.77 & 8.52 & & 7.16 & 12\\
            Li et al.~\cite{Li2019c} & 5.24 & 7.38 & & 3.92 & 5.06 \\          
            Cai et al.~\cite{Cai2020} & 4.6 & 6.55 & & 3.68 & 5.47 \\
            Wang et al.~\cite{Wang2021} & 4.95 & 7.09 & & 3.2 &	4.83 \\
            MS de Arruda et al.~\cite{DeArruda2022} & 4.45 & \textbf{6.18} & & 3.16 & 4.39 \\
            {CenterNet~\cite{Zhou2019}} &  {6.65} & {8.44} & & {1.72} & {2.52} \\  
            CeDiRNet (our) & \textbf{4.43} & 6.33 & & \textbf{1.68} & \textbf{2.61} \\
            \bottomrule
        \end{tabular}
    \end{adjustbox}
    \caption{Results on CARPK and PUCPR+~\cite{Hsieh2017} datasets.}
    \label{tab:carpk_and_pucpr}
\end{table}

Some detections are visualized in Fig.~\ref{fig:sorghum_detections} and~\ref{fig:sorghum_fp_detections} with green and red crosses representing true and false detections, respectively, while ground truth centers are depicted using green dots. Fig.~\ref{fig:sorghum_detections} also shows regressed center-direction fields $\hat{C}_{sin}$ and $\hat{C}_{cos}$ as well as the corresponding direction angle $\phi$. Probability output map from the localization network is also visualized. Depicted detections indicate that CeDiRNet performs really well in majority of cases, while failure cases are often ambiguous examples due to similarity to the sorghum plant (detecting other plants at the edge of the field that are not annotated), due to overlap of multiple plants making it difficult to separate them, or in some cases, due to truncated plants at the edge of the image, which are often not annotated. Truncated and similar plant cases are particularly problematic for this dataset, since they are not always annotated and may be similar to annotated plants, resulting in both conflicting training signal as well as in incorrect detections. In all cases, missing and incorrect detections were the result of incorrect regression of center-direction, while errors in the localization network were never the main cause of issues. Visualization of center-directions also shows that regression of the closest object can be achieved even for pixels farther away from objects thus resulting in greater support for each detection.

\begin{table*}[]
    \centering
    \begin{adjustbox}{width=\textwidth,center}
        \begin{tabular}{cccccccccccccc}
            \toprule
            \multirow{2}{*}{\textit{Method}} & \multirow{2}{*}{\textit{Annotation}} & \multicolumn{3}{c}{Acacia-06} & & \multicolumn{3}{c}{Acacia-12} & & \multicolumn{3}{c}{Oilpalm} \\
            & & prec [\%] & recall [\%] & $F_1$ [\%] & & prec [\%] & recall [\%] & $F_1$ [\%] & & prec [\%] & recall [\%] & $F_1$ [\%] \\
            \midrule
            \midrule
            Faster R-CNN~\cite{Ren2015} & bounding box  & 97.8 & 97.2 & 97.5 & & 94.2 & 96.5 & 95.3 & & 98.9 & 98.3 & 98.5 \\
            FPN~\cite{Lin2016b} & bounding box & 97.6 & 97.4 & 97.6 & & 95.1 & 96.6 & 96.2 & & 98.8 & 97.9 & 98.4 \\
            \midrule
            WSDDN~\cite{Bilen2016} & image level & 77.6 & 70.2 & 71.5 & & 74.3 & 71.8 & 72.9 & & 75.8 & 73.6 & 97.5 \\
            PCL~\cite{Tang2020} & image level & 78.5 & 75.1 & 77.3 & & 79.2 & 76.8 & 78.6 & & 76.4 & 74.7 & 75.9 \\
            C-MIL~\cite{Wan2019} & image level & 87.9 & 82.6 & 86.8 & & 86.5 & 83.9 & 53.0 & & 86.4 & 84.7 & 85.8 \\
            \midrule
            Tong et al.~\cite{Tong2021} & point level & 98.3 & 97.5 & 97.9 & & 95 & 96.7 & 95.8 & & 99.5 & \textbf{99.2} & 99.4  \\
            CeDiRNet (our) & point level & \textbf{98.74} &	\textbf{98.21} & \textbf{98.48} & & \textbf{95.46} & \textbf{97.54} & \textbf{96.49} & & \textbf{99.94} & 99.08 & \textbf{99.51} \\
            \bottomrule
        \end{tabular}
    \end{adjustbox}
    \caption{Results on Acacia-06, Acacia-16 and Oilpalm datasets.}
    \label{tab:plantation_trees}
\end{table*}

Comparing the results with $\tau=5$ and $\tau\ge15$ also indicates that using only 5 pixel error distance for a true positive may be too restrictive and larger $\tau$ would be more appropriate, particularly since most objects in this dataset are always larger than 10 pixels. This is also well shown in detections depicted in Fig.~\ref{fig:sorghum_fp_detections}, where several detections are marked as false positives despite the correct detection of plants.

\paragraph{CARPK and PUCPR+} Results for CARPK and PUCPR+ are shown in Table~\ref{tab:carpk_and_pucpr}. In this table we report only MAE and RMSE metrics since all related literature evaluating on those datasets reports only counting errors. We report localization metrics with $\tau=40$ in Table~\ref{tab:ablatin_focal_and_center}. The results show that the proposed CeDiRNet outperforms the state-of-the-art models on both car datasets. Best performing SOTA model MS de Arruda et al.~\cite{DeArruda2022} is comparable to CeDiRNet on CARPK since SOTA is better in RMSE metric (6.33 versus 6.18, for ours and SOTA respectively), but CeDiRNet is better in MAE metric (4.43 versus 4.45, for ours and SOTA respectively). This discrepancy in MAE and RMSE metrics may indicate fewer errors but a larger variance in CeDiRNet, similar to the results on Sorghum dataset. Nevertheless, CeDiRNet significantly outperforms all remaining models in CARPK dataset{, including CenterNet, which achieved MAE of only 6.65. On PUCPR+ dataset, CeDiRNet slightly outperformed CenterNet,} but significantly outperformed the model by MS de Arruda et al.~\cite{DeArruda2022}, both in MAE (1.68 versus 3.16, for ours and \cite{DeArruda2022} respectively) and RMSE (2.61 and 4.39 for ours and \cite{DeArruda2022} respectively) metrics.

Some examples of detections are also depicted in Fig.~\ref{fig:carpk_detections} for CARPK and in Fig.~\ref{fig:sorghum_fp_detections} for PUCPR+. Both figures also show some difficult occluded examples behind the trees and at the edge of the image, which would be difficult to detect even for human observer. In PUCPR+, some truncated examples are also annotated inconsistently, causing incorrect false detections. Visualization of regressed center-directions for the right image in Fig.~\ref{fig:carpk_detections} also reveals the extent at which center-directions can be regressed. This extends significantly beyond the boundary of the object resulting in a larger support for each center point. Although the regressed values in very distant pixels may be fairly noisy, this does not cause any false detections in this region due to the specific visual appearance that the localization network is tuned to.

\paragraph{Acacia-06/12 and Oil palm} Comparing to the related works that reported results on tree plantation datasets, the proposed center-direction approach outperforms all related methods, as shown in Table~\ref{tab:plantation_trees}. For fair comparison, results are reported using the same $\tau$ as in~\cite{Tong2021}, i.e., using $\tau=40$ for Acacia-06/12 datasets and using $\tau=30$ for Oil palm dataset. CeDiRNet outperforms the state-of-the-art point supervision model~\cite{Tong2021} by around 0.5 - 1 percentage points. Despite a small percentage point difference, this results in over hundreds of fewer errors since Acacia datasets contain around 20,000 objects, while Oil palm dataset contains over 130,000 objects. The proposed center-direction method also outperforms methods that were trained with only image-level annotations as well as bounding box methods, such as Faster R-CNN~\cite{Ren2015} and FPN~\cite{Lin2016b}.

\begin{figure*}
    \centering
    \includegraphics[width=0.75\linewidth]{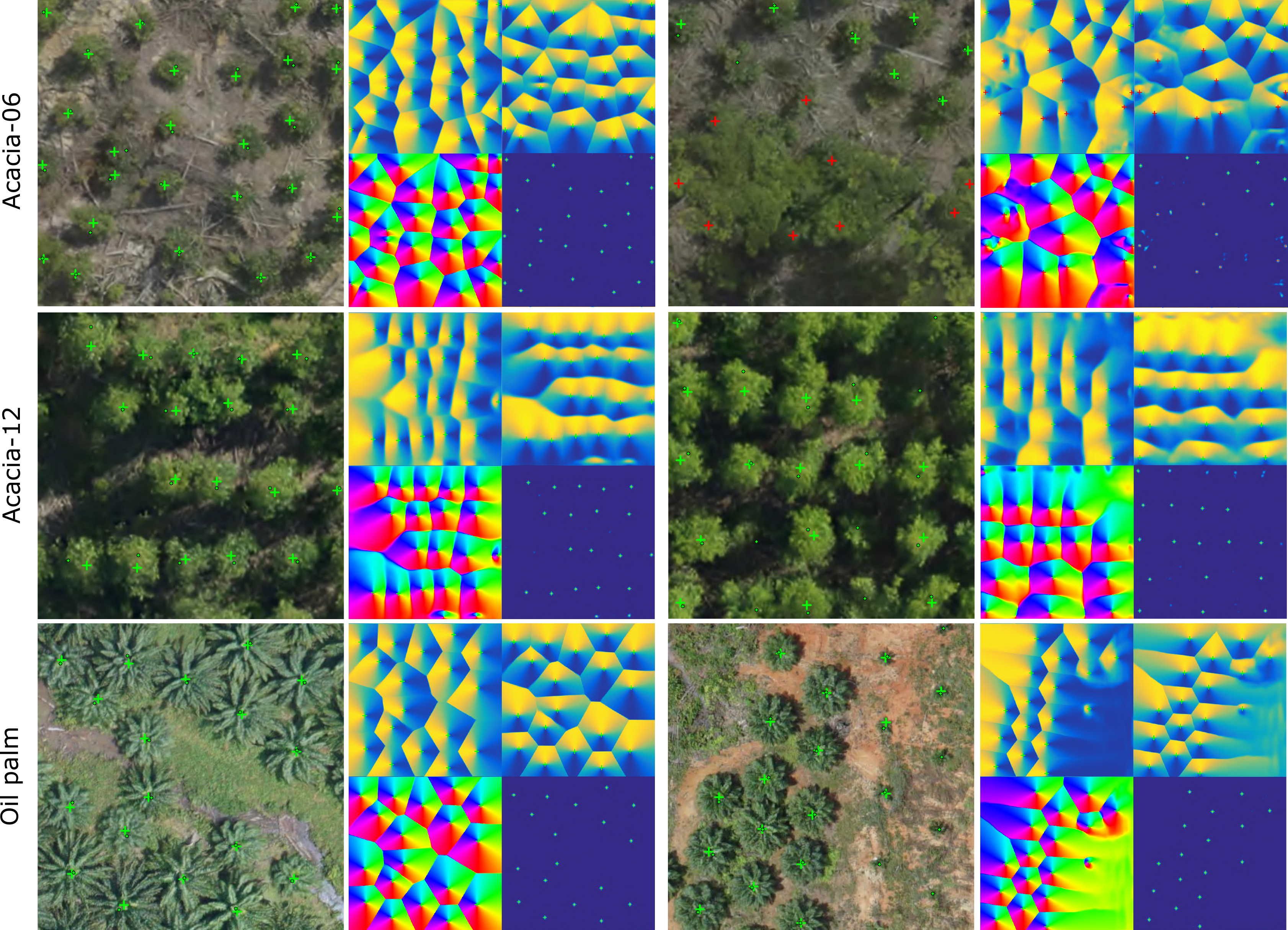}
    \caption{Examples of regressed center-directions and localization detections on Acacia-06, Acacia-12 and Oil palm datasets in top, middle and bottom rows, respectively.}
    \label{fig:plantation_trees_detections}
\end{figure*}

Examples of some correct and incorrect detections are depicted in Fig.~\ref{fig:plantation_trees_detections}. False detections occur mostly on other trees, which are not part of the plantation and are visually similar, while missed detections can arise from smaller/younger trees that are either partly occluded by shade or are visually significantly different from other trees. 

\subsection{Ablation study}

We performed an extensive ablation study that provides further information on our design choices, while also demonstrating additional benefits of the proposed method. 

{
\paragraph{Contribution of background pixels}
In the first experiment, we wanted to verify whether regression of center-directions in pixels located farther away from the center and of the object itself proves useful. In this experiment, the centers are regressed during training only in the pixels that are inside the objects’ bounding boxes. The other (background) pixels are then regressed towards 0 and therefore not used directly for center prediction (as in most related works). We performed this experiment using the CARPK and PUCPR+ datasets, since these two datasets provide bounding boxes for the objects to be detected. In Table~\ref{tab:support_exp}, we can see that the performance decreases compared to the model that also regresses the center-directions in the background pixels. This clearly shows the advantage of regressing center-directions even in the pixels outside the bounding boxes. They do contain information that proves useful for estimating the centers in the second stage of the proposed approach. In addition, we also evaluated a model that completely ignores the pixels outside the bounding boxes, i.e., does not include them in the computation of the loss, which led to even worse results since the inference in the background pixels becomes quite unpredictable.
}

\paragraph{Balancing foreground and background losses}
We also analyzed how sensitive is the proposed weighting scheme for regressing center-directions (Eq.~\ref{eq:weigth_normalization}). The weighting scheme balances the learning of small number of center-directions around objects versus the learning of larger number of center-directions further away from objects. To analyze the sensitivity, we evaluated a model using various cutoff distance threshold values for the weighting scheme. This also included model with $\epsilon=0$, effectively disabling the weighting scheme. All models for this analysis were evaluated on Sorghum \emph{validation set} to avoid overfitting $\epsilon$, as selection of $\epsilon$ for all other experiments was based on this analysis.

Results with various $\epsilon$ values are shown in Table~\ref{tab:ablation_instance_size}. Note that the table reports $2\epsilon$ as this represents the diameter of pixels around the object center instead of a radius. Results demonstrate that well performing models have $2\epsilon$ between 10 and 30 pixels. Smaller values had slightly lower results, while completely disabling the weighting scheme also had a negative effect, with slightly worse localization error. However, $2\epsilon$ values larger than 40 pixels resulted in significantly worse performance, having over 5 - 10 percentage points worse localization error. Prioritizing too large area around the object center may start including too many background pixels that overwhelm the learning around the actual objects.

\begin{figure}
    \centering
    \includegraphics[width=1\linewidth]{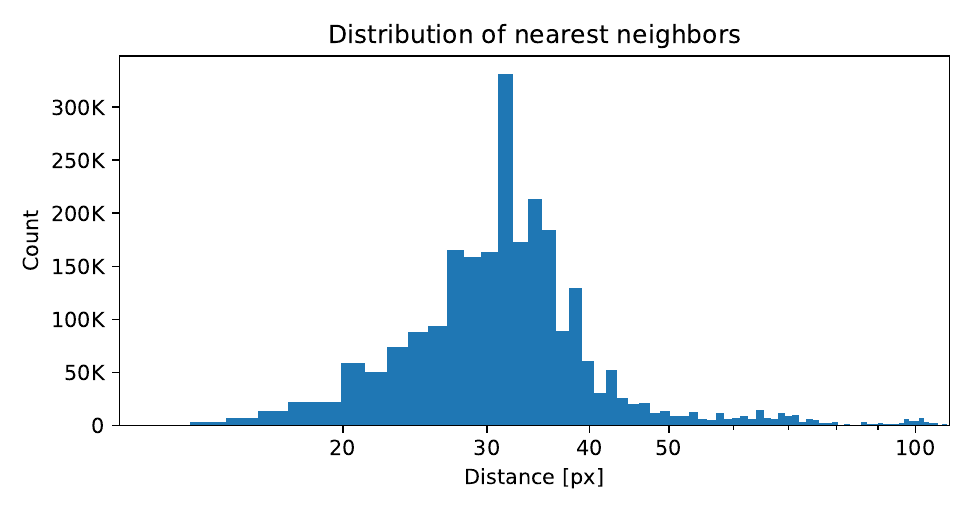}
    \caption{Distribution of distances to nearest neighbor on Sorghum \emph{training} dataset.}
    \label{fig:1NN_distribution}
\end{figure}

Based on this analysis, we fixed $2\epsilon=30$ for all experiments on this dataset as well as on other datasets. Note that $\epsilon$ may also be chosen for each domain specifically based on expected proximity of objects in that domain, which would reduce free hyperparameters for the proposed center-direction learning. A distribution of distances to the nearest object in Sorghum training set is shown in Fig.~\ref{fig:1NN_distribution}. Taking a median would result in $2\epsilon=32$ for this dataset, which is similar to what we used. Moreover, these values could also be set with a more complex schemes, such as ones based on density of objects in each individual image, however, we leave this for a potential future research opportunity.

\begin{table}[]
    \centering
        \begin{adjustbox}{width=1\linewidth,center}
\begin{tabular}{c|ccc|ccc}
            \toprule
            & \multicolumn{3}{c}{CARPK} & \multicolumn{3}{c}{PUCPR+} \\
            & \textit{\makecell[c]{CeDiRNet} } & \textit{\makecell[c]{BB only}} & \textit{no BG} & \textit{\makecell[c]{CeDiRNet} } & \textit{\makecell[c]{BB only}} & \textit{no BG} \\
            \midrule
            prec [\%] & 98.23 & \textbf{98.44} & 81.98 & \textbf{99.57} & 96.81  & 69.27   \\
            recall [\%] & \textbf{94.69} & 94.07 & 84.11 & \textbf{98.41}  & 98.11 & 93.27 \\
            $F_1$ [\%] & \textbf{96.21} & 95.88 & 80.76 & \textbf{98.98} & 97.11 & 71.52  \\
            MAE & \textbf{4.43} & 5.50 & 14.42 & \textbf{1.68} & 1.72 & 23.72 \\
            RMSE & \textbf{6.33} & 8.15 & 19.62 & \textbf{2.61} & 2.75 & 26.22  \\
            \bottomrule
        \end{tabular}   
        \end{adjustbox}
        \caption{{Dense regression of center-directions - \textit{CeDiRNet}: as proposed; \textit{BB only}: only from the object bounding boxes and regressing the other pixels to 0; \textit{no BG}: completely ignoring the background pixels.}}
    \label{tab:support_exp}
\end{table}

\begin{table}[]
    \centering
    \begin{adjustbox}{width=0.9\linewidth,center}
        \begin{tabular}{c|ccc|cc}
            \toprule
            $2\epsilon$  & prec [\%] & recall [\%] & $F_1$ [\%] & MAE & RMSE \\
            \midrule
            \midrule
            0 px & 90.16 & 89.95 & 90.00 & 1.44 & 2.18 \\
            5 px   & 90.47 & 91.01 & 90.67 & 1.52 & 2.43 \\
            10 px   & 91.77 & 91.98 & 91.83 & 1.37 & 2.01 \\
            20 px   & 91.46 & 92.17 & 91.76 & 1.40 & 2.19 \\
            30 px   & 91.05 & 91.29 & 91.12 & 1.33 & 2.02 \\
            40 px   & 83.98 & 84.64 & 84.24 & 1.61 & 2.60 \\
            50 px   & 78.45 & 78.68 & 78.52 & 1.42 & 2.09 \\
            \bottomrule
        \end{tabular}
    \end{adjustbox}
    \caption{Sensitivity analysis of the cutoff distance threshold $\epsilon$ that is used in $W(x_{i,j})$ to emphasize the learning of foreground pixels. Results are reported on Sorghum \emph{validation} set. Note that we report $2\epsilon$ which corresponds to the diameter around the annotation point.}
    \label{tab:ablation_instance_size}
\end{table}

\paragraph{$L_1$ loss for localization network}
For training the localization network, we evaluate the use of $L_1$ loss against the focal loss, which has been commonly used in the literature for learning probability maps of centers. For this experiment, we implement the focal loss by replacing $L_1$ learning loss function $\mathcal{L}_{cent}$ (Eq.~\ref{eq:centernet_loss}) with the focal-loss implementation commonly used in other point-supervision methods~\cite{Wang2021}:
\begin{align}
\label{eq:focal_loss}
    \mathcal{L}_{cent} =& -\dfrac{W_{cent}}{N \cdot M} \sum_{i,j} 
    \begin{cases}
        \Big( 1-\hat{\mathcal{O}}_{cent} \Big)^{\gamma} log \Big(\hat{\mathcal{O}}_{cent} \Big),   & \text{if } \mathcal{O} = 1, \\ 
        A \Big(1-\mathcal{O} \Big)^{\delta} \hat{\mathcal{O}}_{cent}^{\gamma} log \Big(1-\hat{\mathcal{O}}_{cent} \Big), & otherwise
    \end{cases}
\end{align}
where $\hat{\mathcal{O}}_{cent}$ and $\mathcal{O}_{cent}$ are predicted probability and the ground truth of a center, respectively. Loss is applied per-pixel but we omit location $(x_{i,j})$ in both terms for clarity. Hyperparameters $W_{cent}$, $A$, $\delta$ and $\gamma$ were set for best performance by experimenting with four different models based on four different combinations of parameters. As a default version, we set $W_{cent}=1$, $\delta=4$, $\gamma=5$ and $A=\nicefrac{1}{16}$ based on~\cite{Wang2021}, and then experimented with changing $W_{cent}=10$ as well as changing settings $\gamma=2$ and $A=\nicefrac{1}{50}$ (corresponding to $w_{fg}=50$ that was used in our original $L_1$-based loss). Four combinations of focal hyperparameters were optimized together with other two hyperparameters for center-direction regression (weight decay, percent of batch size for hard samples) and the best performing model was selected on the \emph{validation set} for each dataset independently. All focal-loss based models were learned with a learning rate of $0.01$ using the same procedure as used for the $L_1$ loss model, e.g., learning for 200 epochs with a batch size of 768 and using 5000 synthetically generate images. 
Results for the localization network learned with the focal loss are shown in Table~\ref{tab:ablatin_focal_and_center}, where we report the results for all six datasets on the \emph{testing set}. In all cases, the proposed $L_1$ loss models outperformed the focal-loss models. 
\paragraph{Hand-crafted localization network}

We also evaluated an alternative to the localization network, where we used a simple non-learnable hand-crafted network instead of a learnable one. Since the appearance of center-directions changes gradually from -1 to 1 in horizontal and vertical directions, we can replace the 9-layer hourglass network with one layer of 1D convolutions applied in the vertical direction for $C_{sin}$ and the horizontal direction for $C_{cos}$ channels. Instead of learning, the convolution weights can be pre-defined to activate on edges using $[-1, 0, 1]$ filters. We used seven different sizes of kernels, $[3, 9, 15, 21, 31, 51, 65]$, to account for different object scales. Although fairly large convolution weights were used, they are only single dimensional and can be implemented efficiently with significantly smaller computational cost than a normal neural network. Responses to all filters were then summed to form the final response, where local maxima were located. For this experiment, we also optimized hyperparameters for center-direction regression network separately from the previous experiments, since the optimization process is dependent on the final scoring that is provided by the localization network.

\begin{table*}[]
    \centering
    \begin{adjustbox}{width=0.9\textwidth,center}
        \begin{tabular}{c|ccc|ccc|ccc}
            \toprule
            & \textit{$L_1$ loss} & \textit{focal-loss} & \textit{hand-crafted} & \textit{$L_1$ loss} & \textit{focal-loss} & \textit{hand-crafted} & \textit{$L_1$ loss} & \textit{focal-loss} & \textit{hand-crafted}\\            
            \midrule
            &\multicolumn{3}{c}{Sorghum}&\multicolumn{3}{c}{CARPK}&\multicolumn{3}{c}{PUCPR+} \\
            \midrule
            prec [\%] & \textbf{89.76} & 89.49 & 89.50 & \textbf{98.23} & 96.16 & 98.22 & \textbf{99.57} & 99.06 & 98.79 \\
            recall [\%] & 91.03 & 90.99 & \textbf{91.22} & \textbf{94.69} & 94.39 & 94.17 & \textbf{98.41} & 98.37 & 98.02\\
            $F_1$ [\%] & \textbf{90.32} & 90.16 & 90.28 & \textbf{96.21} & 94.74 & 95.94 & \textbf{98.98} & 98.69 & 98.40 \\
            MAE & \textbf{1.78} & 1.83 & 1.86 & \textbf{4.43} & 4.66 & 4.84 & \textbf{1.68} & 1.96 & \textbf{1.68}  \\
            RMSE & \textbf{2.99} & 3.09 & 3.14 & \textbf{6.33} & 6.66 & 7.09 & 2.61 & 2.69 & \textbf{2.51}\\
            \midrule
            &\multicolumn{3}{c}{Acacia-06}&\multicolumn{3}{c}{Acacia-12}&\multicolumn{3}{c}{Oilpalm} \\
            \midrule
            prec [\%] & \textbf{98.74} & 98.66 & 98.60 & 95.46 & 95.05 & \textbf{95.80} & \textbf{99.94} & 99.44 & \textbf{99.94} \\
            recall [\%] & 98.21 & 98.16 & \textbf{98.25} & \textbf{97.54} & 97.36 & 97.41 & 99.08 & \textbf{99.13} & 99.07\\
            $F_1$ [\%] & \textbf{98.48} & 98.41 & 98.42 & 96.49 & 96.19 & \textbf{96.60} & \textbf{99.51} & 99.29 & 99.50 \\
            \bottomrule
        \end{tabular}
    \end{adjustbox}
    \caption{Ablation study on CeDiRNet with different localization network variants.}
    \label{tab:ablatin_focal_and_center}
\end{table*}

Results with hand-crafted localization networks are shown in the last column of Table~\ref{tab:ablatin_focal_and_center}, where we report the results for all six datasets on respective \emph{testing set}. Although hand-crafted network is being out-competed by the learnable network with $L_1$ loss, the results with hand-crafted network are still better than other state-of-the-art models on all datasets except for CARPK. The hand-crafted network even slightly outperforms learnable localization model on Acacia-12 dataset in both $F_1$ and precision ($F_1$ and precision of 96.60\% and 95.80\%, respectively, vs 96.49\% and 95.46\%), while in PUCPR+ it outperforms learnable localization model in counting error based on counting metrics (MAE and RMSE of 1.68 and 2.51, respectively, vs 1.68 and 2.61) but not in localization error ($F_1$ of 98.40\% vs 98.98\%).

Such good performance compared to other state-of-the-art models demonstrates that regression of center-directions exposes information on the location of objects in a manner that can be extracted by a fairly simple process of 1D convolutions. Moreover, this can be achieved at a low computational cost since 1D convolution can be implemented efficiently with significantly smaller cost than regular convolutional network. A solution with hand-crafted localization network may be particularly suitable for applications where computational cost plays an important factor, while in applications where computational cost is less of an issue, slight performance gain can be achieved by using the learnable network. 

\paragraph{Domain-specific fine-tuning of the localization network}

Next, we tested whether a generic localization network could be further improved by fine-tuning it on a specific domain. In this experiment, we used a pre-trained generic localization network for initialization and then learned the localization model simultaneously with the center-direction regression network using a real output data of center-direction regressions instead of the synthetic one. Synthetic data was used only when learning the pre-trained localization model.

\begin{table}[]
    \centering
    \begin{adjustbox}{width=1\textwidth,center}
        \begin{tabular}{cc|ccc|cc}
            \toprule
            synthetic & \makecell[c]{domain}  & prec [\%] & recall [\%] & $F_1$ [\%] & MAE & RMSE \\
            \midrule
            \midrule
                       &   & 89.50 & \textbf{91.22} & 90.28 & 1.86 & 3.14 \\
            \checkmark &   & \textbf{89.76} & 91.03 & \textbf{90.32} & \textbf{1.78} & \textbf{2.99} \\
            \checkmark & \checkmark  & 89.26 & 90.90 & 90.00 & 1.84 & 3.03 \\
            \bottomrule
        \end{tabular}
    \end{adjustbox}
    \caption{Ablation study on using different training data types for the localization networks on Sorghum testing dataset. Results are shown for: (a) non-learnable hand-crafted network in the first row, (b) generic network trained only on synthetic data in the second row, and (c) additional domain-specific fine-tuning applied to the generic network in the third row. }
    \label{tab:ablation_localization_data}
\end{table}

The results on the domain-specific fine-tuning are shown in Table~\ref{tab:ablation_localization_data}, where we report the results for Sorghum dataset on the \emph{testing set}. Comparison of the second and  the third row that show learnable localization models without and with fine-tuning, respectively, indicates that the domain-specific fine-tuning did not improve performance in any metric. \textcolor{blue}{Observing a performance decline during fine-tuning may seem counterintuitive, given the expectation of added benefits from domain-specific information. This may be attributed to the risk of overfitting to specific visual appearances in real training data stemming from limited diversity in center-directions (e.g., their count and dispersion) and the absence of augmentation (e.g., synthetic occlusions and noise). These factors pose more significant constraints than training on synthetic data where randomly generated locations ensure uniqueness for each batch. This suggests that the visual appearance of center-directions remains sufficiently constrained, facilitating effective use of synthetic data for training the localization network and simplifying application to diverse domains, where only the center-direction regression model needs to be learned.}

Compared to non-learnable hand-crafted model, the domain-specific fine-tuning is still slightly better in terms of count error, but in terms of the localization error it performed worse in both precision and recall. Nevertheless, the differences between all three models are fairly small, and domain-specific fine-tuning also outperforms the best state-of-the-art model reported on this dataset. 

\paragraph{{Robustness analysis of the localization network}}

\begin{figure*}
    \centering
    \includegraphics[width=0.4\linewidth]{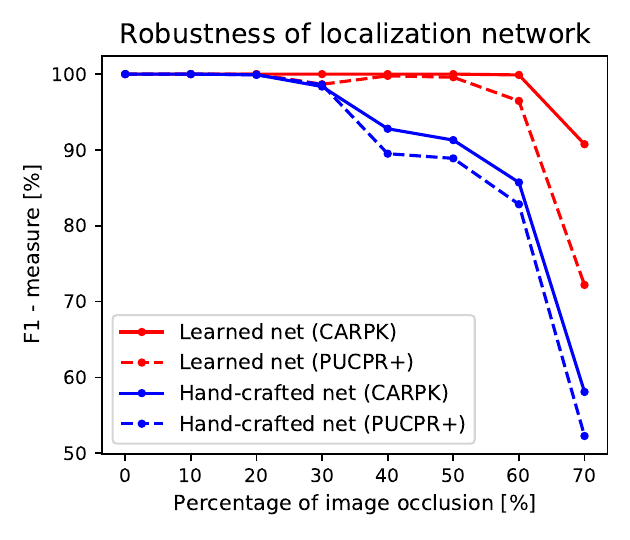}~~\includegraphics[width=0.38\linewidth]{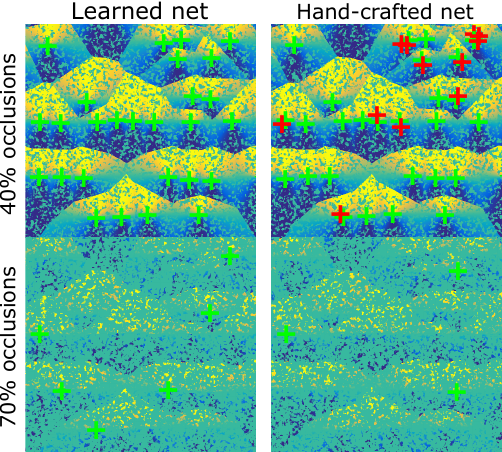}
    \vspace{10pt}
    \includegraphics[width=0.79\linewidth]{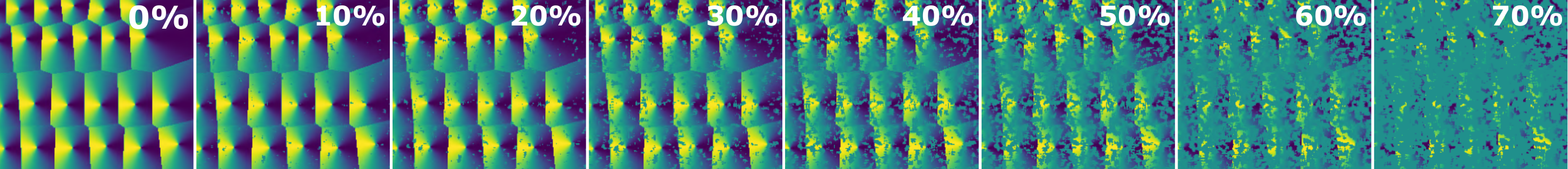}

    \caption{{Performance of localization network at various levels of occlusion noise on the left, with one example on the right showing cosine of center-direction at 0.4 and 0.7 occlusion noise for learned and hand-crafted network, and various degrees of occlusion noise at the bottom.} }
    \label{fig:robusness_exp}
\end{figure*}

{We performed additional analysis of the localization network to assess the robustness to failures of the regressed center-directions. We performed an experiment by artificially corrupting center-direction values with various levels of occlusion and running corrupted images through the localization network. We created an occlusion noise by uniformly sampling values for each pixel, then blurring them with the Gaussian kernel using $\sigma=5$ and finally thresholding them to get small occluded regions. By varying the threshold, different percentages of occluded pixels were obtained, as can be seen at the bottom in Figure~\ref{fig:robusness_exp}. 
}

{We applied occlusion masks to the groundtruth data of CARPK and PUCPR+ datasets. Results are shown in Figure~\ref{fig:robusness_exp}. We can observe robust detection even in the presence of serious occlusions. Both networks are fairly robust to up to 30\% of image occlusion. The learned localization network has proven to be significantly more resilient than the hand-crafted network as it is able to maintain performance even at an extreme level of occlusion of up to 60\%, as depicted in Figure~\ref{fig:robusness_exp}. Possible explanation for such robustness can be found in augmentation with random Gaussian noise applied during the training of the localization network.}

\paragraph{{Inference running times}}
{
In Table~\ref{tab:runtime}, we also report the inference runtime analysis. Running times were calculated on images of $1280\times736$ pixels in size using a single NVIDIA GeForce RTX 2080 Ti GPU and Intel Xeon Silver 4214 CPU. We report averaged measurements repeated ten times, each time performing a warm-up with 100 runs before measuring actual runtime. We report results for CeDiRNet model with both localization networks and use CenterNet model for comparison. 
Regression of center-directions in CeDiRNet required an additional 20\% of the total time compared to the CenterNet. This can be contributed to the FPN decoder used by CeDiRNet, since CenterNet does not use it. The localization network in CeDiRNet then took additional 10 ms for the learned network, while hand-crafted network took only 2 ms. In both cases the time needed for post-processing was almost negligible. The proposed method is therefore quite fast, operating at around 13 - 15~FPS on HD images.
}
\section{Conclusion}
\label{sec:conclusion}

In this work we presented a novel method for counting and localization based on point supervision by using a novel center-direction approach termed CeDiRNet. We proposed to split the problem into first performing dense regression of center-directions and then performing localization of centers from densely regressed center-direction fields. Formulating center-directions as a re-parametrization of sine and cosine values enabled detection of object's center location with a larger support from multiple surrounding pixels. At the same time, it also pawed the way for using generic non-complex model for localization of centers, which can be not only trained from synthetic data but also does not require any domain-specific fine-tuning. This facilitates application to different domains by simply learning the center-direction regression model. 

\begin{table*}[]
    \centering
    \begin{adjustbox}{width=0.65\textwidth,center}
    \begin{tabular}{cc|ccc}
    \toprule
        \textit{Method} & \textit{Localization network} & \textit{Regression} & \textit{Localization} & \textit{Post-processing}  \\
    \midrule
        \multirow{2}{*}{CeDiRNet (our)} & Learned & $64.7\pm0.3$ ms & $10.6\pm0.2$ ms & $0.6\pm0.0$ ms \\
         & Hand-crafted & $64.4\pm0.6$ ms & $1.9\pm0.0$ ms & $0.5\pm0.0$ ms \\
    \midrule        
        CenterNet & / & \multicolumn{2}{c}{$53.6\pm0.3$ ms} &  $1.3\pm0.1$ ms \\
    \bottomrule
    \end{tabular}
    \end{adjustbox}
    \caption{{Inference runtimes reported on an image of size $1280\times736$ using a single GPU.}}
\label{tab:runtime}

\end{table*}

We have demonstrated on six different datasets for counting and localization in the remote sensing domain that CeDiRNet outperforms all existing state-of-the-art models for counting and localization that are trained with point supervision. Regression of center-directions is thus shown to be a viable solution for learning dense prediction of center points from a set of sparse annotations, which in other approaches is addressed with focal loss and fine-tuning of their hyperparameters for each domain. With center-directions this problem can be confined only to the learning of one generic localization network, which can be trained once on synthetic data, while in each domain we simply learn dense regression of center-directions. Even applying focal loss to the learning of our localization network is shown as not needed. This also simplifies learning of the localization network, and allows to use a simple local maxima for locating the final centers in the localization heatmap without the need to apply additional post-processing heuristics.

The proposed CeDiRNet also has a high applicative value. This stems directly from the concept of pointing towards the centers, which enables: (a) training with a low-cost point annotations, (b) learning of only center-direction regression for different domains and using generic localization network trained on synthetic data, and (c) simplified learning with fewer hyperparameters and no additional post-processing heuristics. 

{Despite demonstrated state-of-the-art performance on several remote sensing domains there are certain limitations of the proposed method. CeDiRNet method can be sensitive to certain types of occlusions, particularly when two or more objects are positioned such that they have the same center point and surrounding pixels from different objects point to the same location thus making impossible for the localization network to distinguishes them (e.g., a large ship carrying a smaller one). An additional limitation can also be found in dealing with highly crowded scenes, where small objects are densely packed together, resulting in a small area around each object center. This limits the number of pixels that point to a specific center and can result in blending the neighboring direction vectors, making localization network difficult to distinguish between objects. This limitation could be addressed by increasing the resolution of both high-level feature maps as well as resolution of the output directions thus providing more pixels that would point towards the center. 
}

The concept of regressing center-directions is also not limited to counting or localization and may be applied to other tasks as well. In future work, we plan on extending CeDiRNet beyond point-supervised localization by applying it to instance segmentation task. In instance segmentation, background pixels could be learned as zero-magnitude center-directions, thus serving dual purpose of encoding segmentation mask and providing direction to the centers. Each instance could then be determined from a group of regressed center-directions that all point towards the same center. Since the same model can also be used for point-supervised localization, this provides an elegant approach for learning with mixed-supervision that would provide a way to reduce the annotation effort, i.e., enable using instance annotation for a smaller subset of images and point annotation for others. {Our preliminary results also showed promising results when adding perturbations to groundtruth centers, which may merit more research focus in the future.} Finally, the proposed concept of regressing center-directions is also agnostic to any specific backbone architecture and could be implemented with any modern deep learning architecture for dense prediction. While we used ResNet-FPN in our work, we already found promising results with ConvNext and transformer-based backbones, but we leave further optimization for both regression and localization networks as future research opportunities.

\section*{Acknowledgments}
This work was in part supported by the ARRS research project J2-3169 (MV4.0) and J2-4457 (RTFM) as well as by research programme P2-0214.

\bibliographystyle{elsarticle-num} 
\bibliography{library_cedirnet}

\end{document}